\documentclass[10pt,twocolumn,letterpaper]{article}

\usepackage[pagenumbers]{cvpr} 

\usepackage[dvipsnames]{xcolor}
\usepackage{color, colortbl}
\usepackage{pifont}  
\usepackage{mathtools}
\usepackage{algorithm,algorithmicx,algpseudocode}
\usepackage{multirow}

\newcommand{\methodname}{NPCD}

\definecolor{mylightgreen}{RGB}{51,153,102}
\definecolor{mydarkgreen}{RGB}{17,121,74}

\definecolor{mygreenblue}{RGB}{47,111,112}
\definecolor{mylightblue}{RGB}{172,207,208} 

\definecolor{myyellow}{RGB}{215,179,18} 
\definecolor{myorange}{RGB}{215,136,18}

\definecolor{mylightred}{RGB}{192,72,72} 
\definecolor{mydarkred}{RGB}{158,28,28} 

\definecolor{mylightgray}{RGB}{238,238,238} 
\definecolor{mymedgray}{RGB}{202,204,206} 
\definecolor{mydarkgray}{RGB}{42,44,46}

\definecolor{mymaxturbo}{RGB}{161,17,1}
\definecolor{myminturbo}{RGB}{57,42,119}

\definecolor{appearance}{RGB}{103,169,207}
\definecolor{positions}{RGB}{239,138,98}

\definecolor{rowhighlight}{gray}{0.9}
\colorlet{bgcolor}{mylightgray}
\colorlet{poscolor}{mydarkgreen}
\colorlet{negcolor}{mydarkred}
\colorlet{outputscalingcolor}{blue}
\definecolor{draftcolor}{RGB}{0,73,95}

\definecolor{cvprblue}{rgb}{0.21,0.49,0.74}

\def\rot#1{\rotatebox{90}{#1}}

\newcommand{\mn}{\ding{55}}

\newcommand\blfootnote[1]{%
  \begingroup
  \renewcommand\thefootnote{}\footnote{#1}%
  \addtocounter{footnote}{-1}%
  \endgroup
}

\newcommand{\realnumbers}{\mathbb{R}}

\newcommand{\pointthreed}{\mathbf{p}}

\newcommand{\pointfeature}{\mathbf{f}}
\newcommand{\pointfeaturedim}{D}
\newcommand{\shadingpoint}{\mathbf{q}}

\DeclareMathOperator*{\argmin}{arg\,min}
\newcommand{\norm}[1]{\left\lVert#1\right\rVert}
\newcommand{\object}{O}

\newcommand{\view}{V}

\newcommand{\neuralpointcloud}{\mathcal{P}}
\newcommand{\neuralpointcloudnum}[1]{\neuralpointcloud_{#1}}
\newcommand{\posmatrix}{\mathbf{P}}
\newcommand{\fmatrix}{\mathbf{F}}
\newcommand{\image}{\mathbf{I}}

\newcommand{\numobjects}{N}
\newcommand{\numviews}{K}
\newcommand{\numpoints}{M}
\newcommand{\nerfcolor}{\mathbf{c}}
\newcommand{\density}{\sigma}
\newcommand{\localmlp}{F_{\phi}}
\newcommand{\colormlp}{G_{\psi}}
\newcommand{\shapemlp}{H_{\gamma}}
\newcommand{\loss}{\mathcal{L}}

\usepackage[pagebackref,breaklinks,colorlinks,citecolor=cvprblue]{hyperref}
\usepackage[accsupp]{axessibility} 

\def\paperID{xxx} 
\def\confName{CVPR\xspace}
\def\confYear{2024\xspace}

\title{Neural Point Cloud Diffusion\\for Disentangled 3D Shape and Appearance Generation}

\author{
Philipp Schr\"oppel\textsuperscript{1}\hspace*{-0.1cm}\\
\and
Christopher Wewer\textsuperscript{2}\\
\and
Jan Eric Lenssen\textsuperscript{2}\\
\and
Eddy Ilg\textsuperscript{3}\\
\and
Thomas Brox\textsuperscript{1}\\
\and
\textsuperscript{1}University of Freiburg, Freiburg, Germany\\
\textsuperscript{2}Max Planck Institute for Informatics, Saarland Informatics Campus, Germany\\
\textsuperscript{3}Saarland University, Saarland Informatics Campus, Germany\\
}

\begin{document}
\makeatletter
\let\@oldmaketitle\@maketitle
\renewcommand{\@maketitle}{\@oldmaketitle
%
%
\vspace*{-1.2cm}
\begin{center}
        \centering
      \includegraphics[width=1\linewidth]{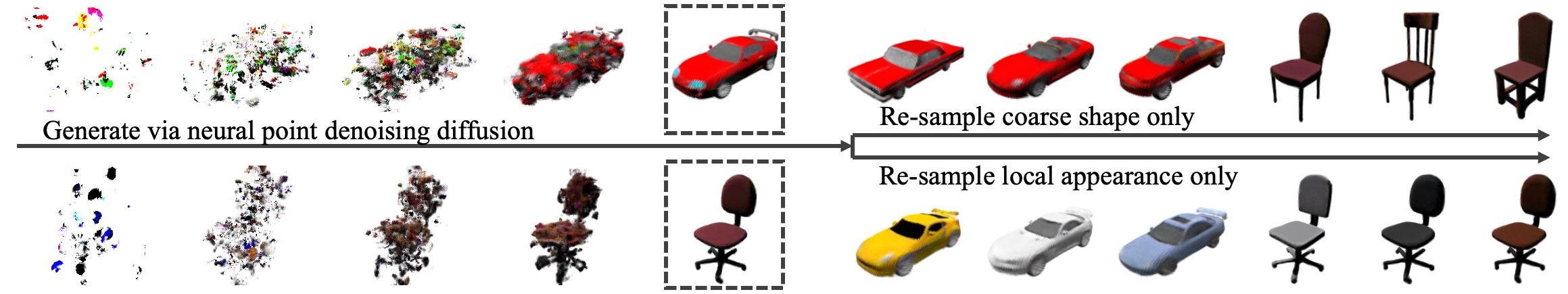}
        \captionsetup{aboveskip=0pt}
        \captionof{figure}{We present a method to model 3D radiance field distributions using neural point denoising diffusion~\textbf{(left)}. Since our representation disentangles coarse object shape from local appearance, we can sample from the individual distributions separately~\textbf{(right)}.
        }
        \label{fig:teaser}
\end{center}
\vspace*{1mm}
 }
\makeatother
\maketitle

\begin{abstract}
Controllable generation of 3D assets is important for many practical applications like content creation in movies, games and engineering, as well as in AR/VR.
Recently, diffusion models have shown remarkable results in generation quality of 3D objects.
However, none of the existing models enable disentangled generation to control the shape and appearance separately.
For the first time, we present a suitable representation for 3D diffusion models to enable such disentanglement 
by introducing a hybrid point cloud and neural radiance field approach. We model a diffusion process over point positions jointly with a high-dimensional feature space for a local density and radiance decoder. 
While the point positions represent the coarse shape of the object, the point features allow modeling the geometry and appearance details.
This disentanglement enables us to sample both independently and therefore to control both separately. Our approach sets a new state of the art in generation compared to previous disentanglement-capable methods by reduced FID scores of \mbox{30-90\%} and is on-par with other non-disentanglement-capable state-of-the art methods. 
\end{abstract}
    
\section{Introduction}
\label{sec:Introduction}

3D assets are used in many practical applications, ranging from engineering to movies and computer games, and will become even more important in virtual spaces and virtual telepresence that will be enabled by AR/VR technology in the future. However, manually creating such assets is a labor-intensive and costly task that requires expert skills. It is even more important that such content cannot just be generated but that the generation can also be controlled to obtain the desired outcome. With the impressive image generation capabilities of diffusion models~\cite{Ho2020,diffbeatgan,dalle,Rombach2022,imagegen,controlnet}, it is appealing to consider such models.  Generally, the extension to 3D is still limited and not straightforward~\cite{Nichol2022,Shue2023, Erkoc2023}. Even more so, none of the current diffusion models allow for \emph{disentangled} generation of shape and appearance and controlling them separately. 

The general challenge for 3D diffusion models lies in selecting the right 3D representation. 
One track of work explores diffusion models to generate 3D point clouds~\cite{Cai2020, Luo2021, Zhou2021, Zeng2022, Nichol2022}. While such methods are able to generate the sparse point clouds well, they are not able to model dense geometry or appearance.
Another track uses implicit representations~\cite{Erkoc2023,Jun2023,Dupont2022}, triplanes~\cite{Shue2023,Chen2023} or voxel grids~\cite{Muller2023} to define the geometry and appearance continuously for each coordinate in a volume that encloses the object. The downside of all of these representations is that they do not provide disentanglement of shape and appearance. 
The reason for this limitation is the missing invariance of a single factor of variation to changes in others: 
Global neural field representations, for example, model shape and appearance in joint parameters.
Thus, one of these factors cannot be changed independently. 
Voxel grids or triplanes provide limited invariance of appearance variables to voxel-sized shifts but fail to be invariant to sub-voxel shifts or more complex, non-rigid sub-voxel deformation.

In contrast, we propose a method that enables individual generation of shape and appearance by introducing a hybrid approach that consists of a neural point cloud hosting a continuous radiance field. The  point cloud explicitly disentangles coarse object shape from appearance. Feature vectors model the geometry and appearance of \emph{local parts}~\cite{deepls} and, while the point positions determine \emph{where} a part is, the point features describe \emph{how} the details of a part look like. Notably, the point positions can undergo complex deformations without requiring changes in point features. With this representation, we are able to control the generation of both aspects separately, as illustrated in~\cref{fig:teaser}.  

To establish our model, we first train a category-level Point-NeRF autodecoder~\cite{Xu2022} by sharing its weights across many instances of ShapeNet~\cite{Sitzmann2019} or PhotoShape~\cite{Park2018} objects. The obtained neural point clouds then serve as a dataset to train a diffusion model that learns to denoise the point positions and features simultaneously. Different to previous diffusion models, our model operates on high-dimensional latent spaces. 
In summary, our contributions are: 
\begin{enumerate} 
\item We propose the first approach for object generation that leverages a hybrid approach consisting of a neural point cloud combined with a neural renderer and a diffusion model that operates in a high-dimensional latent space. 
\item We identify many-to-one mappings as a crucial obstacle when applying denoising diffusion to autodecoded, high-dimensional latent spaces and present effective regularization schemes to overcome this issue. 
\item We show that our approach is capable of successfully disentangling geometry and appearance by allowing to control them separately and that the generation quality our approach significantly outperforms the previous methods GRAF~\cite{schwarz2020graf} and Disentangled3D~\cite{tewari2022disentangled3d} by a large margin, while being on-par in generation quality with state-of-the-art methods incapable of disentangling.  
\end{enumerate}

\section{Related work}
\label{sec:RelatedWork}

\paragraph{Disentangled generation.}
Disentangled generation of shape and appearance has been studied in multiple works and is commonly achieved by modeling distinct architecture parts and latent codes~\cite{zhu2018visual,niemeyer2020differentiable,jang2021codenerf}. GRAF~\cite{schwarz2020graf,niemeyer2021giraffe} presented the first generative model for radiance fields that allows to separately control both factors. Disentangled3D (D3D)~\cite{tewari2022disentangled3d} provides a more explicit disentanglement by leveraging a canonical volume and deformation. In contrast to ours, none of the approaches uses diffusion models or point clouds. We can show that our diffusion model clearly outperforms these previous GAN-based methods.

\paragraph{Probabilistic diffusion models.}
In recent years, diffusion models~\cite{Sohldickstein2015, Song2019} have emerged as a successful class of generative models. They first define a forward diffusion process in the form of a Markov chain, which gradually transforms the data distribution to a simple known distribution. A model is then trained to reverse this process, which then allows to sample from the learned data distribution. DDPM~\cite{Ho2020} proposes a diffusion model formulation with various simplifications that enable high quality image synthesis. Follow-up works improved on synthesis quality via refinements of the architecture and sampling procedure to even outperform state-of-the-art generative adversarial networks~\cite{diffbeatgan, Karras2022}.
In order to scale diffusion models to high resolutions, LDM~\cite{Rombach2022} moves the diffusion process from the image to a latent space with smaller spatial dimensions. In this work, we adopt the DDPM diffusion formulation and apply it to 3D objects on a high-dimensional latent space in the form of neural point clouds.

\paragraph{NeRF and Point-NeRF.}
NeRF~\cite{Mildenhall2020} represents geometry and appearance as a radiance and density field that can be volumetrically rendered to photorealistic images. Point-NeRF~\cite{Xu2022} extends NeRF to a parameterization by a point cloud that is obtained from MVSNet~\cite{mvsnet}. This reduces ambiguity and allows for a much faster reconstruction.   
The Point-NeRF MLPs are originally trained on a single scene. In this work, we train them jointly on many objects as a category-level autodecoder~\cite{wewer23simnp}. In addition, we regularize the neural point cloud features to be optimally suitable for the diffusion model, which we use to generate objects. 

\paragraph{3D diffusion.} There are currently two trends of applying diffusion to 3D:
(1) Test-time distillation using large pre-trained image generators~\cite{Poole2023, Lin2023, Melas2023, Liu2023, Zhou2023} and (2) diffusion models on datasets of 3D models~\cite{Cai2020, Luo2021, Zhou2021, Zeng2022, Nichol2022, Dupont2022, Jun2023, Muller2023, Chen2023, Anciukevicius2023, Chan2022, Shue2023}.
Our approach can be assigned to the second category, and therefore we focus on this direction in the following.

\paragraph{Diffusion on point clouds.}
Unlike alternative 3D representations, such as dense voxel grids or meshes, point clouds are sparse, unlimited to pre-defined topologies, and flexible w.r.t. modifications.
Therefore, there are several recent works combining these advantages with the generative power of diffusion models~\cite{Cai2020, Luo2021, Zhou2021, Zeng2022, Nichol2022}.
Except for differences in network architecture and the relation to score matching or diffusion models, first approaches~\cite{Cai2020, Luo2021, Zhou2021} all define the diffusion process directly on 3D point coordinates.
In contrast to that, LION~\cite{Zeng2022} applies the idea of LDM~\cite{Rombach2022} to point clouds by denoising latent codes of a hierarchical VAE. However, with only a single dimension, their latent codes are not very expressive and very low dimensional. 
Following the great success of text-to-image generation, Point-E~\cite{Nichol2022} trains a transformer-based architecture for generation of RGB point clouds conditioned on complex prompts. 
Unlike all of these approaches, our method generates point clouds with high-dimensional features encoding detailed shape and appearance.

\paragraph{Diffusion for 3D object shape and appearance.}
Previous approaches for object shape and appearance synthesis opt for other 3D representations.
Functa~\cite{Dupont2022} and Shap-E~\cite{Jun2023} generate the weights of implicit (neural) representations such as radiance fields or signed distance functions.
DiffRF~\cite{Muller2023} uses a 3D-UNet to denoise explicit voxel grids storing density and color.
The usual training pipeline for these approaches involves first fitting 3D representations to a dataset of multi-view images.
Recently, SSDNeRF~\cite{Chen2023} showed improvements by optimizing the diffusion model and individual NeRFs for each training object in a joint single stage.
RenderDiffusion~\cite{Anciukevicius2023} avoids the question about single- or two-stage training by defining the diffusion process again in image space, but it appends the triplane representation from EG3D~\cite{Chan2022} to the usual diffusion architecture for 3D view conditioning.
In contrast to implicit representations, voxel grids and triplanes, our neural point clouds enable the disentanglement of shape and appearance.

\section{Method}
\label{sec:Method}

\begin{figure}[t]
        \centering
        \includegraphics[width=1\linewidth]{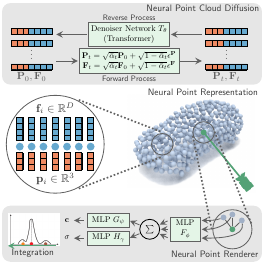}
        \caption{ \textbf{Overview of neural point cloud diffusion (NCPD).} In the \textbf{center} we have a neural point cloud representation, where each point has a position (\textcolor{positions}{$\blacksquare$}) and an appearance feature (\textcolor{appearance}{$\blacksquare$}). The neural point cloud can be generated with a diffusion model \textbf{(top)} and can be rendered via ray integration \textbf{(bottom)}. }
        \label{fig:method_fig}
        \vspace{-0.4cm}
\end{figure}

In this section, we describe Neural Point Cloud Diffusion (NPCD), our generative model for 3D shape and appearance via diffusion on neural point clouds. An overview of the method is shown in \cref{fig:method_fig}. At the center of our method is an autodecoder with a neural point representation for the latent codes, which is further described in \cref{sec:PointNeRFTraining}. We discuss characteristics of autodecoder schemes in \cref{sec:ManyToOne} and provide regularization schemes that enable denoising diffusion on the feature space. Subsequently, in \cref{sec:DiffusionTraining} we then present a diffusion model to denoise the neural point positions and features. After the diffusion model is trained, we can sample 3D shape and appearance independently from each other as described in \cref{sec:MethodDisentanledGeneration}. \blfootnote{We provide code for our method at {{\url{https://github.com/lmb-freiburg/neural-point-cloud-diffusion}}}.}

\subsection{Category-Level Point-NeRF Autodecoder}
\label{sec:PointNeRFTraining}
We begin by outlining our representation as an extension of Point-NeRF~\cite{Xu2022}. A single object is represented  by a neural point cloud $\neuralpointcloud=\{(\pointthreed_1, \pointfeature_1), ..., (\pointthreed_\numpoints, \pointfeature_\numpoints)\}=(\posmatrix,\fmatrix)$ where each 3D point $i\in \{1,...,M\}$ with position $\pointthreed_i\in \realnumbers^3$ is associated with a neural feature $\pointfeature_i\in\realnumbers^\pointfeaturedim$. We denote the full matrix of point positions as $\posmatrix \in \realnumbers^{M\times 3}$ and the full matrix of features as $\fmatrix \in \realnumbers^{M\times \pointfeaturedim}$. In contrast to PointNeRF, we assume the point positions $\posmatrix$ to be given for training objects and $\fmatrix$ to be manually initialized before optimization. We explore different initialization strategies in \cref{sec:NeuralPointcloudInitialization}. The neural point representation can be rendered from arbitrary views, as described in the following.

\vspace{-0.2cm}
\paragraph{Volume rendering.} 
To render a pixel of an image we follow the Point-NeRF~\cite{Xu2022} procedure. Given camera parameters, we march rays through the scene and sample shading points $\shadingpoint$ along the ray. For each shading point, the features $\mathbf{f}$ of the neighboring points $\mathbf{p}$ of the neural point cloud are first aggregated to a shading point feature $\pointfeature_\shadingpoint$ via a multi-layer perceptron (MLP) $\localmlp$ and a weighted combination based on inverse distances:
\begin{equation}
    \pointfeature_\shadingpoint = \sum\limits_{i=1}\frac{w_i \localmlp(\pointfeature_i, \shadingpoint - \pointthreed_i)}{\sum\limits_{i=1}w_i}, \text{where } w_i=\frac{1}{||\shadingpoint - \pointthreed_i||_2}.
\end{equation}
This feature is then mapped mapped to a color $\nerfcolor$ and density $\density$ by separate MLPs $\shapemlp$ and $\colormlp$:
\begin{equation}
    \nerfcolor = \colormlp(\pointfeature_\shadingpoint) \qquad \density = \shapemlp(\pointfeature_\shadingpoint)
\end{equation}
The obtained radiances are then numerically integrated to the pixel color as described in NeRF~\cite{Mildenhall2020}. Given camera parameters $\mathbf{v}$, we denote the full rendering function that renders an image as $R^\mathbf{v}_{\phi,\psi, \gamma}(\posmatrix, \fmatrix)$. 

\vspace{-0.2cm}
\paragraph{Optimization.} Optimization is done on a dataset of $\numobjects$ objects $\object_1, ..., \object_\numobjects$. Each object $\object_j$ consists of a neural point cloud $\neuralpointcloud_j = (\posmatrix_j, \fmatrix_j)$ and $\numviews$ views $\view_{j1}, ..., \view_{j\numviews}$. Each view $\view_{jk}=(\image_{jk},\mathbf{v}_{jk})$ consists of a ground truth image $\image_{jk}$ and corresponding camera parameters $\mathbf{v}_{jk}$.
The optimization objective is to jointly find the point features $\mathbf{F}$ and network parameters $\phi, \psi,\gamma$ that minimize the image reconstruction error for all views of all objects:
\begin{equation}
    \hat{\fmatrix}, ´\hat{\phi}, ´\hat{\psi}, \hat{\gamma} =  \argmin_{\mathbf{F}, ´\phi, \psi, \gamma} \sum_{j,k}\mathcal{L}\left(R^{\mathbf{v}_{jk}}_{\phi,\psi, \gamma}(\posmatrix_j, \fmatrix_j), \mathbf{I}_{jk}\right) \textnormal{,}
    \label{eq:PointNeRFLoss}
\end{equation}
with $\loss$ being the mean squared error between rendered and ground truth pixel colors.
In contrast to Point-NeRF, we share parameters $\phi, \psi, \gamma$ of the rendering MLPs over all objects as an autodecoder with $(\posmatrix_j, \fmatrix_j)$ as latent codes.

\subsection{Autodecoding for diffusion}
\label{sec:ManyToOne}
The objective given above in \cref{eq:PointNeRFLoss} describes an underconstrained optimization problem. We found that, without further regularization, many-to-one mappings between features $\fmatrix$ and renderings $R^\mathbf{v}_{\theta,\psi, \phi}(\posmatrix, \fmatrix)$ emerge. Thus, there are multiple possible point features $\mathbf{f}_i$ representing the same local appearance information. We support this hypothesis with an empirical verification in \cref{sec:NeuralPointcloudRegularization} and \cref{tab:Cosine}.  The many-to-one mappings pose a challenge for the denoising model, as it is trained to produce point estimates of the features $\mathbf{f}_i$ that are in that case ambiguous. 

Most existing latent diffusion methods circumvent this issue by using an autoencoder~\cite{Rombach2022, Zeng2022} instead of optimizing representations via backpropagation. Since encoder networks are functions by design, and thus assigning each input value only one output, they do not produce many-to-one mappings between latent representation and output. However, we argue that the autodecoder principle is preferred in many situations, since it does not require an encoder (which is difficult to design for representations like neural points) and often leads to higher quality.

Therefore, we present and analyze a list of strategies to eliminate many-to-one mappings from autodecoder formulations, which are outlined in the following paragraphs.

\vspace{-0.2cm}
\paragraph{Zero initialization.} The first simple, albeit effective strategy is to initialize features $\mathbf{F}$ with zero instead of randomly sampled values. We can show that this is very effective and encourages convergence to the same minimum. 

\vspace{-0.2cm}
\paragraph{Total variation regularization (TV).} Inspired by TV regularization on triplanes~\cite{Shue2023}, we design a TV regularization baseline for our neural point clouds by adding
\begin{equation}
    \mathcal{L}_{TV}(\mathbf{F}) = \lambda_{TV} \sum_{i=1}^M\sum_{n\in \mathcal{V}(i)} \frac{\norm{\mathbf{f}_i  -\mathbf{f}_n}_1}{\norm{\mathbf{p}_i  -\mathbf{p}_n}_2} \,,
    \label{eq:PointNeRFTVLoss}
\end{equation}
with weighting $\lambda_{TV}$ to the objective in \cref{eq:PointNeRFLoss} for each object, where $\mathcal{V}(i)$ is a local neigborhood of points around the point with index $i$. Intuitively, it encourages that neighboring point features are varying only slightly.

\vspace{-0.2cm}
\paragraph{Variational autodecoder (KL).} We introduce a variational autodecoder by storing vectors of means $\mathbf{\mu}_i$ and isotropic variances $\mathbf{\Sigma}_i$ instead of features $\mathbf{f}_i$ for each point. For rendering, we obtain the features $\mathbf{f}_i$ by sampling from the corresponding Gaussians using the reparameterization trick~\cite{Kingma2014}. Additionally, we add a KL divergence loss
\begin{equation}
    \mathcal{L}_{KL}(\{\mu_i, \mathbf{\Sigma}_i\}_{i=1}^M) = \lambda_{KL} \sum_{i=1}^M KL(\mathcal{N}(\mathbf{\mu}_i, \mathbf{\Sigma}_i),\mathcal{N}(\mathbf{0}, \mathbf{I}_\pointfeaturedim)) \,,
    \label{eq:PointNeRFKLLoss}
\end{equation}
for each object with weighting $\lambda_{KL}$, to minimize the evidence lower bound~\cite{Kingma2014}. Intuitively, the consequences from this regularization are twofold. First, the latent space is regularized to follow a unit Gaussian distribution, reducing many-to-one mappings in the representation. Second, the decoder learns to be more robust to small changes in $\mathbf{f}$ due to the sampling procedure. Note that our diffusion model regresses $\mathbf{\mu}_i$, but we write $\mathbf{f}_i$ in the following for simplicity.

\subsection{Neural point cloud diffusion}
\label{sec:DiffusionTraining}
In this section, we describe our diffusion model for neural point cloud representations. As input, we assume a set of optimized representations $\{\neuralpointcloudnum{j}\}_{j=1}^N$ from the first stage. The diffusion model learns the distribution of these representations, which allows us to generate neural point clouds.

\paragraph{Denoising diffusion background.} 
Diffusion models~\cite{Sohldickstein2015, Song2019} learn the distribution $q(\mathbf{x})$ of data $\mathbf{x}$ by defining a forward diffusion process in form of a Markov chain with steps $q(\mathbf{x}_t|\mathbf{x}_{t-1})$ that gradually transform the data distribution into a simple known distribution. A model $p_\theta(\mathbf{x}_{t-1}|\mathbf{x}_{t})$ with parameters $\theta$ is then trained to approximate the steps $q(\mathbf{x}_{t-1}|\mathbf{x}_{t})$ in the Markov chain of the reverse process, which allows to sample from the learnt distribution.

In case of Gaussian diffusion processes, the forward diffusion process gradually replaces the data with Gaussian noise following a noise schedule $\beta_1, .., \beta_T$:
\begin{equation}
q(\mathbf{x}_t|\mathbf{x}_{t-1}) \coloneqq \mathcal{N}(\mathbf{x}_t;\sqrt{1-\beta_t}\mathbf{x}_{t-1},\beta_t \mathbf{I})\,. 
\label{eq:forwardprocess}
\end{equation}
Further, it is possible to sample $\mathbf{x}_t$ directly from $\mathbf{x}_0$:
\begin{equation}
q(\mathbf{x}_t|\mathbf{x}_{0}) = \mathcal{N}(\mathbf{x}_t;\sqrt{\bar\alpha_t}\mathbf{x}_{0},(1-\bar\alpha_t) \mathbf{I})\,,
\label{eq:forwardprocessskip}
\end{equation}
with $\alpha_t \coloneqq 1-\beta_t$ and $\bar\alpha_t \coloneqq \prod_{s=1}^t \alpha_s$.
For small enough step sizes in the noise schedule, the steps in the Markov chain of the reverse process can be approximated with Gaussian distributions: 
\begin{align}
p_\theta(\mathbf{x_T}) &= \mathcal{N}(\mathbf{x}_{T}; \mathbf{0}, \mathbf{I}), \\ p_\theta(\mathbf{x}_{t-1}|\mathbf{x}_{t}) &= \mathcal{N}(\mathbf{x}_{t-1};\mu_\theta(\mathbf{x}_{t}, t),\Sigma_\theta(\mathbf{x}_{t}, t))\,. 
\label{eq:reverseprocess}
\end{align}
The objective hence is to learn $\mu_\theta(\mathbf{x}_{t}, t)$ and $\Sigma_\theta(\mathbf{x}_{t}, t)$. 
DDPM~\cite{Ho2020} proposes a diffusion model formulation with various simplifications. Specifically, DDPM suggests to fix $\Sigma_\theta(\mathbf{x}_{t}, t)$ and the noise schedule $\beta_t$ to time-dependent constants. Further, DDPM reparametrizes \cref{eq:forwardprocessskip} to 
\begin{equation}
\mathbf{x}_t(\mathbf{x}_0, \mathbf{\epsilon}) = \sqrt{\bar\alpha_t}\mathbf{x}_0 + \sqrt{1-\bar\alpha_t}\mathbf{\epsilon} \text{ with } \mathbf{\epsilon} \sim \mathcal{N}(\mathbf{0}, \mathbf{I})
\label{eq:forwardprocessskip2}
\end{equation}
and trains the model to directly predict $\mathbf{\epsilon}_\theta(\mathbf{x}_t, t) $, from which $\mu_\theta(\mathbf{x}_{t}, t)$ can be computed.

\paragraph{Neural point cloud diffusion.} Given the background in DDPM, we turn to describing our neural point cloud diffusion. Conceptually, the neural points clouds $\neuralpointcloud=(\posmatrix,\fmatrix)$ take the place of data points $\mathbf{x}_0$ in the above DDPM diffusion model formulation.

The distribution of both modalities, point positions $\posmatrix$ and appearance features $\fmatrix$, is learnt jointly. During training, we sample Gaussian noise $\mathbf{\epsilon}^{\posmatrix}$ for all point positions and $\mathbf{\epsilon}^{\fmatrix}$ for all point features and use it to obtain the noised neural point cloud $\neuralpointcloud_t = (\posmatrix_t, \fmatrix_t)$ at a specific timestep $t$ in the diffusion process via \cref{eq:forwardprocessskip2}. The denoiser network $T_\theta((\posmatrix_t, \fmatrix_t), t)=(\mathbf{\epsilon}^{\posmatrix}_\theta, \mathbf{\epsilon}^{\fmatrix}_\theta)$ takes the noised neural point cloud and timestep as input and estimates the noise $\mathbf{\epsilon}^{\posmatrix}_\theta$ and $\mathbf{\epsilon}^{\fmatrix}_\theta$ that was applied to the points and features. The network is optimized by minimizing the average mean squared error on both noise vectors.

\paragraph{Denoiser architecture.} As architecture for the denoiser network, we use a Transformer~\cite{Nichol2022, Vaswani2017, Peebles2022DiT}. As input, the transformer receives $\numpoints+1$ tokens, one token per point plus one additional token encoding $t$. Point tokens are obtained by concatenating the point position and feature of each point in the noisy point cloud and projecting them with a linear layer. Similarly, the $t$ token is obtained by projection with its own linear layer onto the same dimensionality. After encoding, all tokens are processed by transformer layers, including self-attention and MLPs. Finally, the resulting output tokens corresponding to the $\numpoints$ points are projected back to the dimensions of the input point positions and features and interpreted as noise predictions $\mathbf{\epsilon}^{\posmatrix}_\theta$ and $\mathbf{\epsilon}^{\fmatrix}_\theta$.

\subsection{Disentangled generation}
\label{sec:MethodDisentanledGeneration}
Given a trained NPCD model, we can naively sample from the joint distribution $p(\posmatrix, \fmatrix)$ of point positions and features by sampling positions and features from a unit Gaussian distribution and using the transformer for iterative denoising (c.f. \cref{fig:teaser} for a visualization). To achieve individual generation of shape and appearance, one needs to sample from conditional distributions $p(\posmatrix | \fmatrix)$ or $p(\fmatrix | \posmatrix)$ instead.

For appearance-only sampling from $p(\fmatrix | \posmatrix)$ with given point positions $\posmatrix_0$, we obtain the initial noisy neural point cloud by sampling $\fmatrix_T$ from a unit Gaussian distribution and computing $\posmatrix_T$ from $\posmatrix_0$ via the forward process in \cref{eq:forwardprocessskip}.

Throughout the diffusion process, we update the point features $\fmatrix_{t-1}$ from the denoiser outputs according to the reverse diffusion process in \cref{eq:reverseprocess}, but update the point positions $\posmatrix_{t-1}$ from the given $\posmatrix_0$ via the forward diffusion process in \cref{eq:forwardprocessskip} instead of the denoiser outputs. 
Sampling from $p(\posmatrix | \fmatrix)$ is done analogously. Further details are provided in the supplementals. 

Intuitively, this is comparable to masked image inpainting, using an approach similar to RePaint~\cite{Lugmayr2022}. Instead of masking image parts, we mask one modality of our representation. Note that this is enabled by the neural point representation, which allows to disentangle the variables for coarse shape and local appearance. 
\section{Experiments}
\label{sec:DisentangledGeneration}
In this section, we provide experimental results for the presented NPCD method. We begin by introducing the experimental setup in \cref{sec:exp_setup} and used metrics in \cref{sec:metrics}. Then, we evaluate the main contribution of our method in Sec.~\ref{sec:disentangled_generation}, \ie the distentangled generation of coarse geometry and appearance. We compare against previous generative approaches that allow disentangled generation, namely GRAF~\cite{schwarz2020graf} and Disentangled3D~\cite{tewari2022disentangled3d}, and show our superior generation quality. Next, we compare against recent diffusion models without disentangling capabilities in Sec.~\ref{sec:sota_comparison}. Here, we compare against DiffRF~\cite{Muller2023}, Functa~\cite{Dupont2022} and  SSDNeRF~\cite{Chen2023}. Many existing 3D generative models model only shape but not appearance. Thus, to complement existing comparisons, we also provide a shape-only comparison in Sec.~\ref{sec:shape_only_comparison}.
Finally, we analyze many-to-one mappings due to auto-decoded features as a problem for diffusion models and propose regularization methods as effective countermeasures in Sec.~\ref{sec:Analyses}. 

\subsection{Datasets and experimental setup}
\label{sec:exp_setup}

\paragraph{Data.} We use the cars and chairs categories of the ShapeNet SRN dataset~\cite{Chang2015, Sitzmann2019}. The cars split contains $2{,}458$ training objects and $704$ test objects, while the chairs split contains $4{,}611$ training and $1{,}317$ test objects. We use the original renderings with 50 views per training object and 251 views per test object. The images have a resolution of 128x128 pixels. For all training objects, the poses are sampled randomly from a sphere. For all test objects the poses follow a spiral on the upper hemisphere. We extract point clouds with 30k points from the mesh and subsample them to 512 points with farthest point sampling. 

Additionally, we use the PhotoShape Chairs dataset~\cite{Park2018}. The dataset contains $15{,}576$ objects and features more realistic textures on top of ShapeNet meshes. We use the same test split as DiffRF~\cite{Muller2023} with $1,552$ objects. From the remaining objects, we randomly select $2,480$ objects for training. We use the same renderings as DiffRF~\cite{Muller2023}, which consist of 200 views per object on an Archimedean spiral. We use a resolution of 128x128 pixels and the same point clouds with 512 points as for SRN chairs.

\begin{figure*}[t!]
        \centering
        \subfloat[]{
            \begin{tabular}{@{}cc@{}}
              \raisebox{.5\height}{\includegraphics[width=0.06\linewidth]{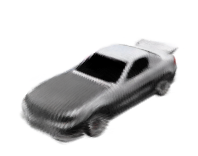}} &
              \includegraphics[width=0.4\linewidth]{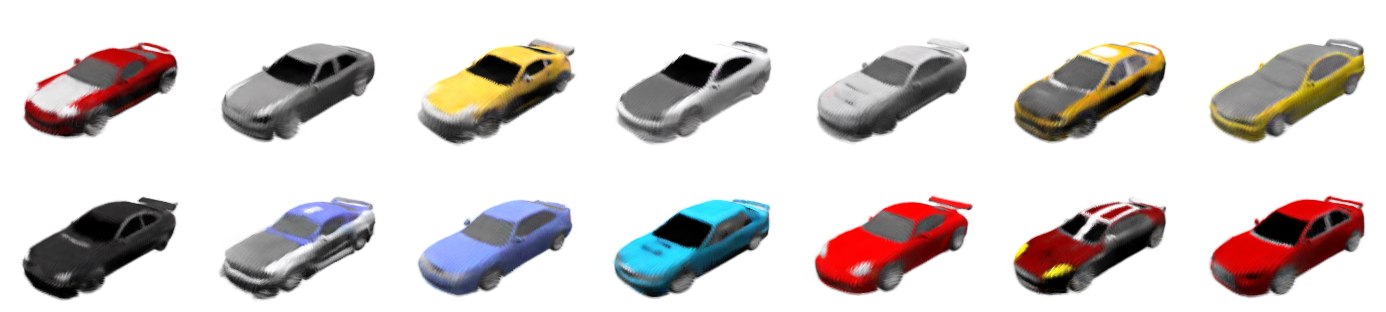} \\
              \raisebox{.5\height}{\includegraphics[width=0.06\linewidth]{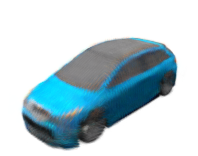}} &
              \includegraphics[width=0.4\linewidth]{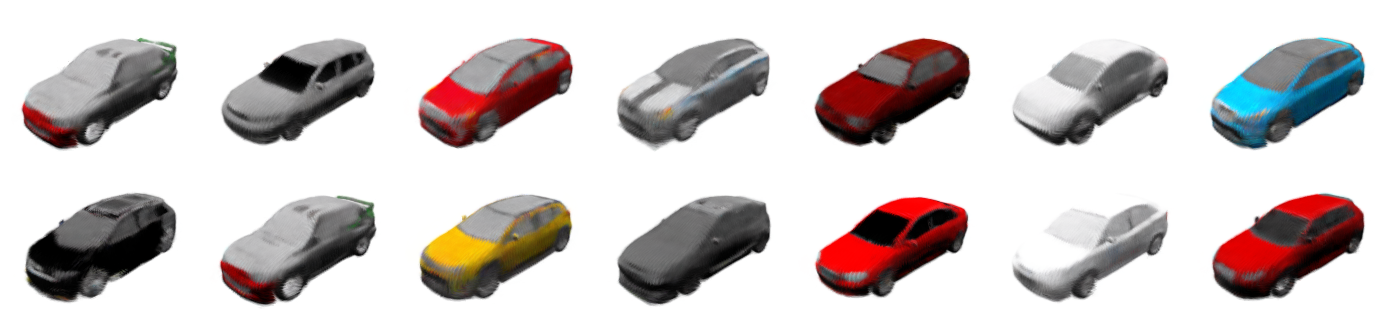} \\ 
              \raisebox{.5\height}{\includegraphics[width=0.06\linewidth]{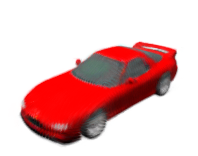}} &
              \includegraphics[width=0.4\linewidth]{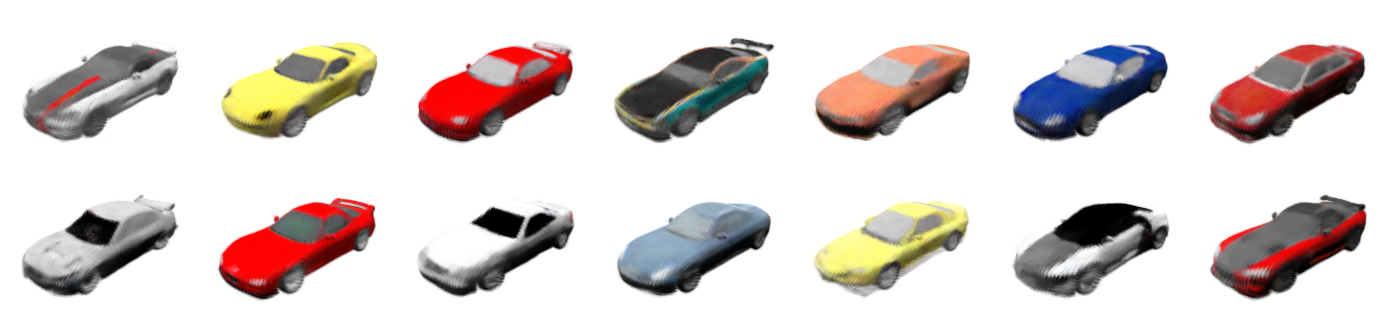} \\ \midrule
              \raisebox{.5\height}{\includegraphics[width=0.06\linewidth]{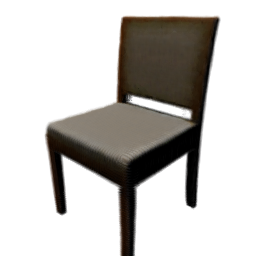}} &
              \includegraphics[width=0.4\linewidth]{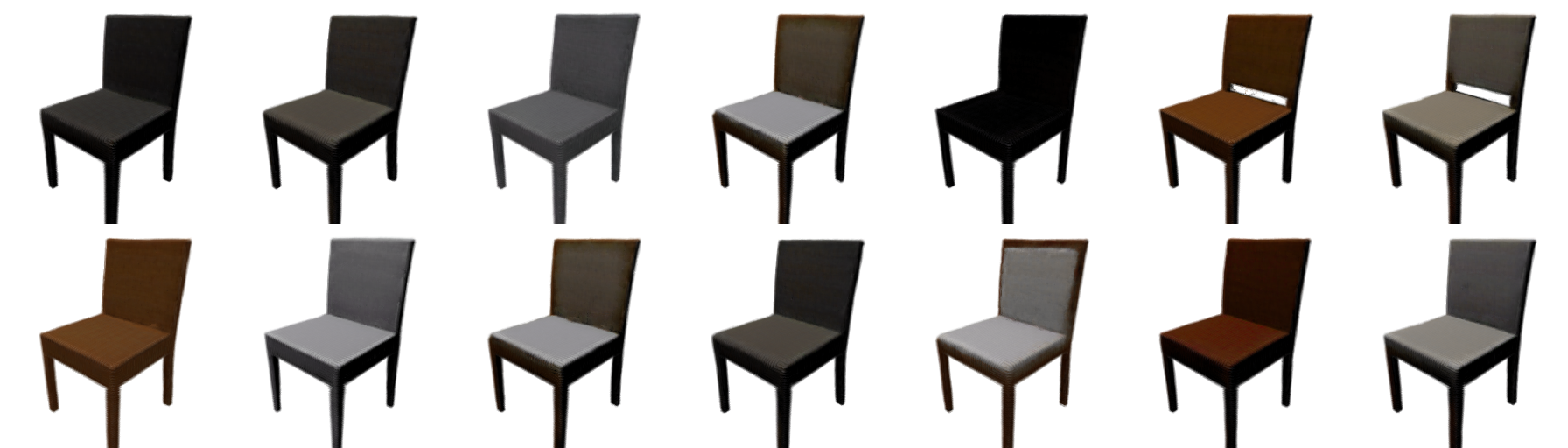} \\
              \raisebox{.5\height}{\includegraphics[width=0.06\linewidth]{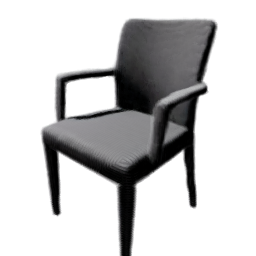}} &
              \includegraphics[width=0.4\linewidth]{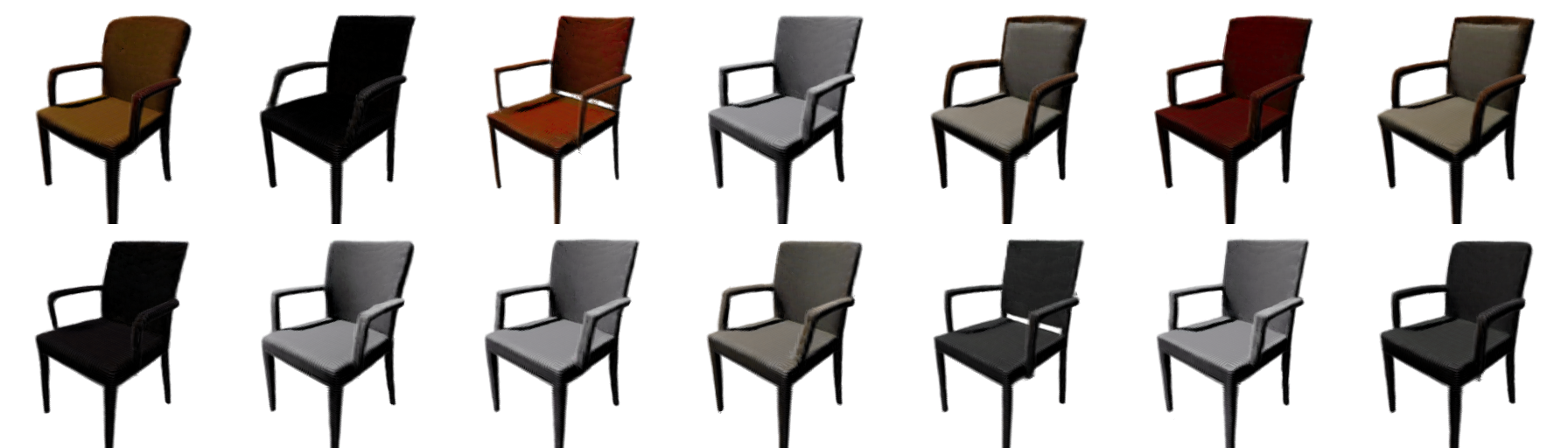}  \\ \midrule
              \raisebox{.5\height}{\includegraphics[width=0.06\linewidth]{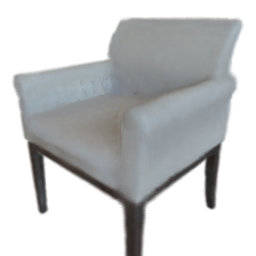}} &
              \includegraphics[width=0.4\linewidth]{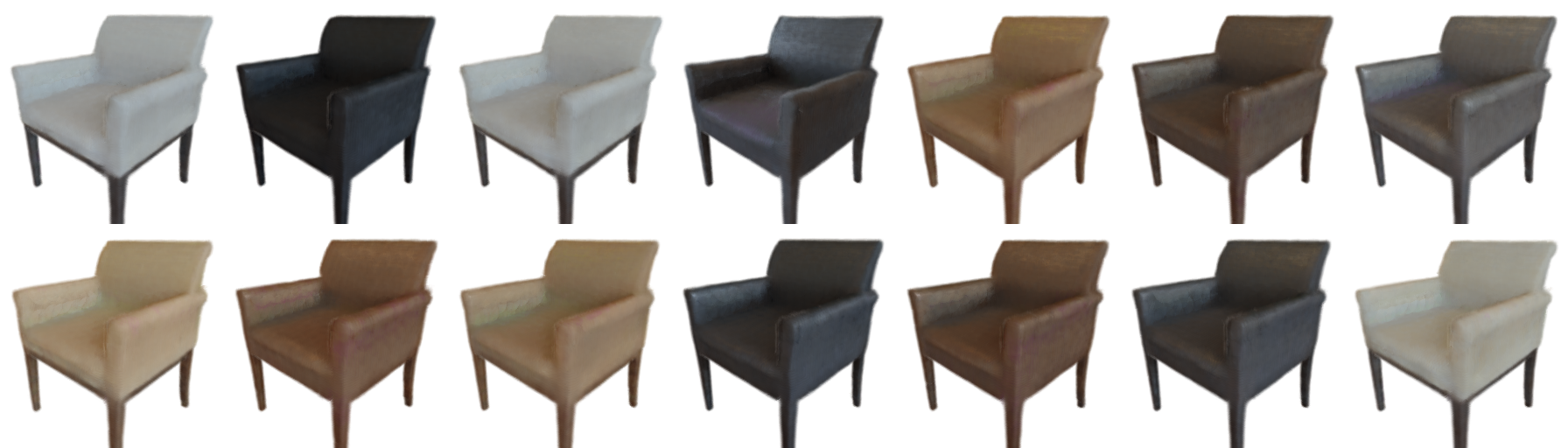} \\
              \raisebox{.5\height}{\includegraphics[width=0.06\linewidth]{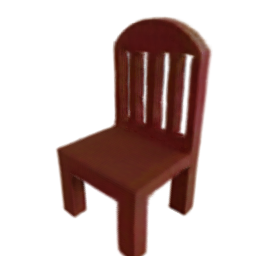}} &
              \includegraphics[width=0.4\linewidth]{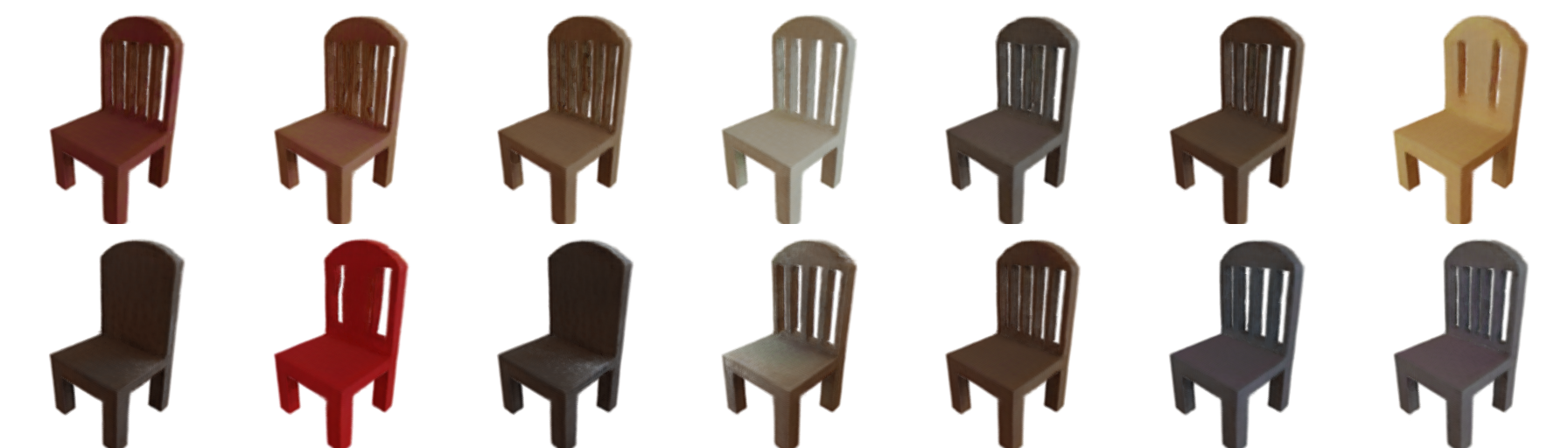} 
          \end{tabular}
          \label{subfig:DisentanglementAppearanceOnly}
        }\unskip \vrule
        \subfloat[]{
            \begin{tabular}{@{}cc@{}}
                \raisebox{.5\height}{\includegraphics[width=0.06\linewidth]{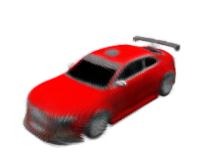}} &
              \includegraphics[width=0.4\linewidth]{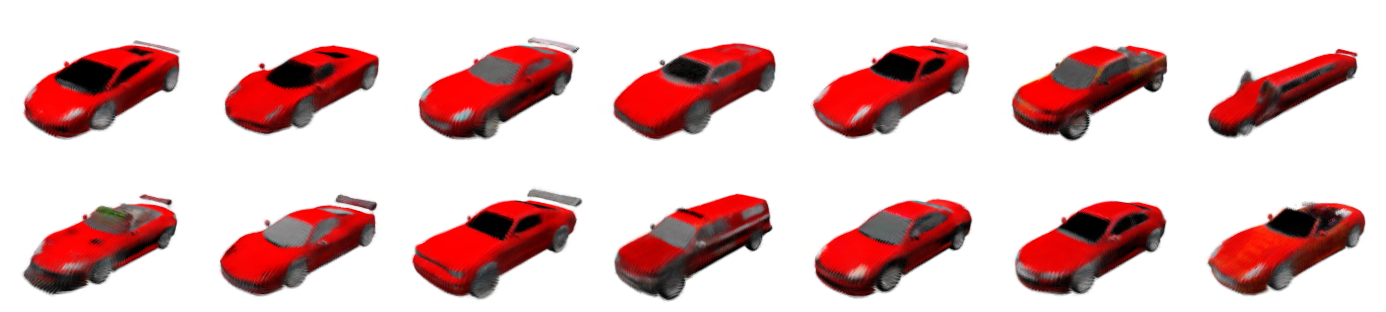} \\
                \raisebox{.5\height}{\includegraphics[width=0.06\linewidth]{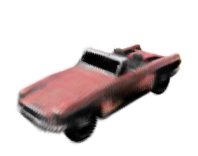}} &
              \includegraphics[width=0.4\linewidth]{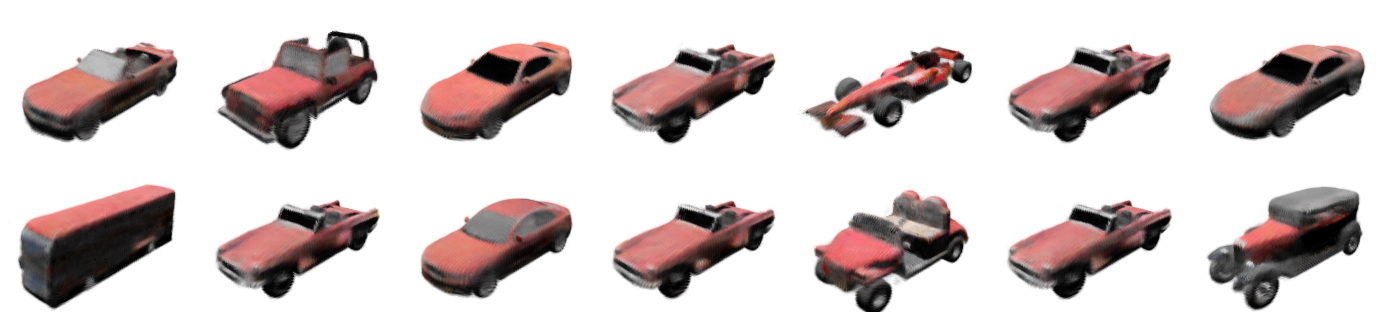} \\
              \raisebox{.5\height}{\includegraphics[width=0.06\linewidth]{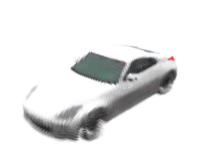}} &
              \includegraphics[width=0.4\linewidth]{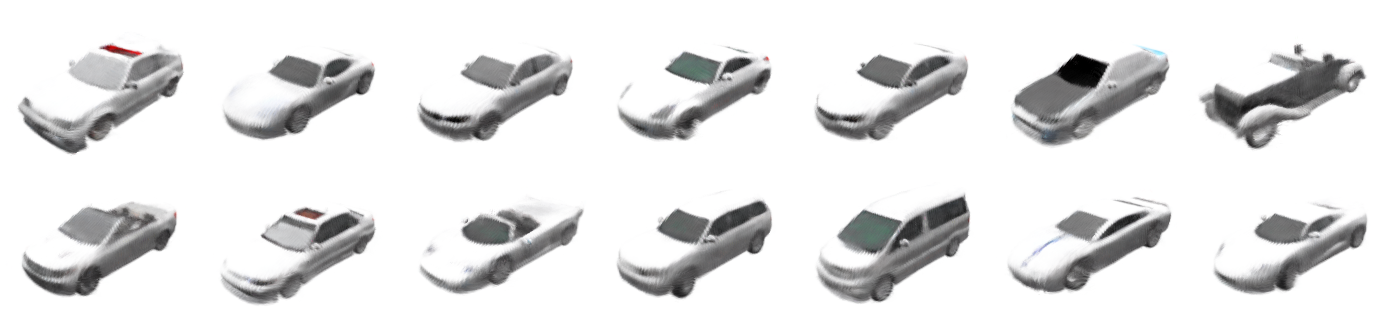} \\ \midrule
              \raisebox{.5\height}{\includegraphics[width=0.06\linewidth]{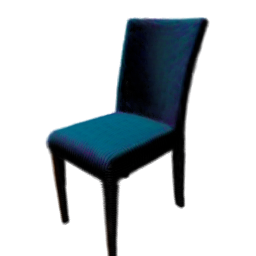}} &
              \includegraphics[width=0.4\linewidth]{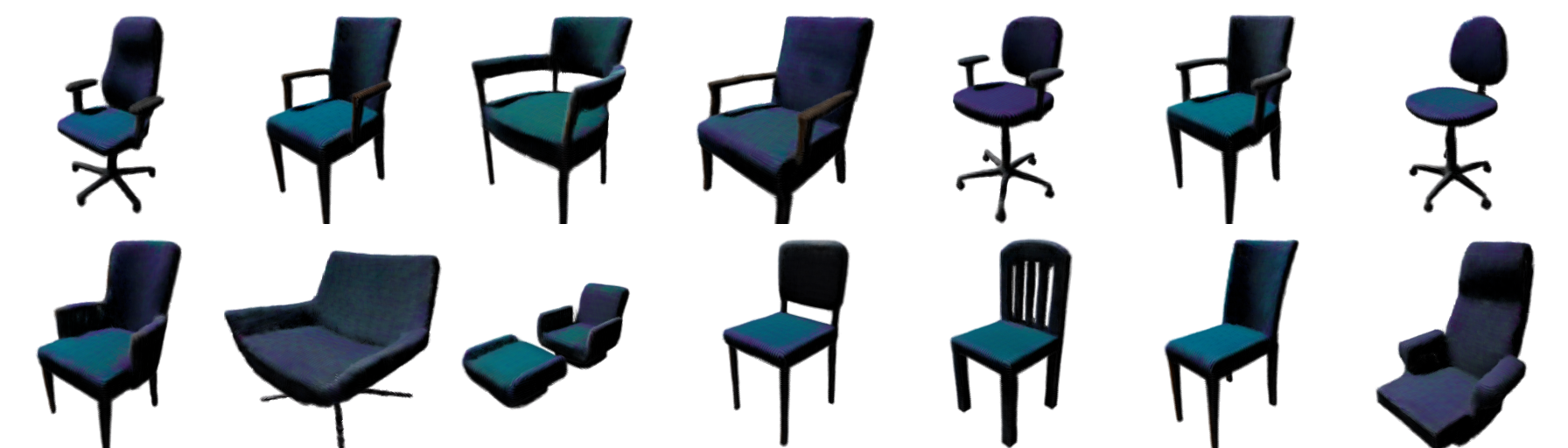}  \\
              \raisebox{.5\height}{\includegraphics[width=0.06\linewidth]{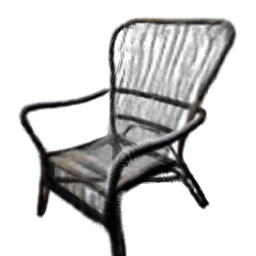}} &
              \includegraphics[width=0.4\linewidth]{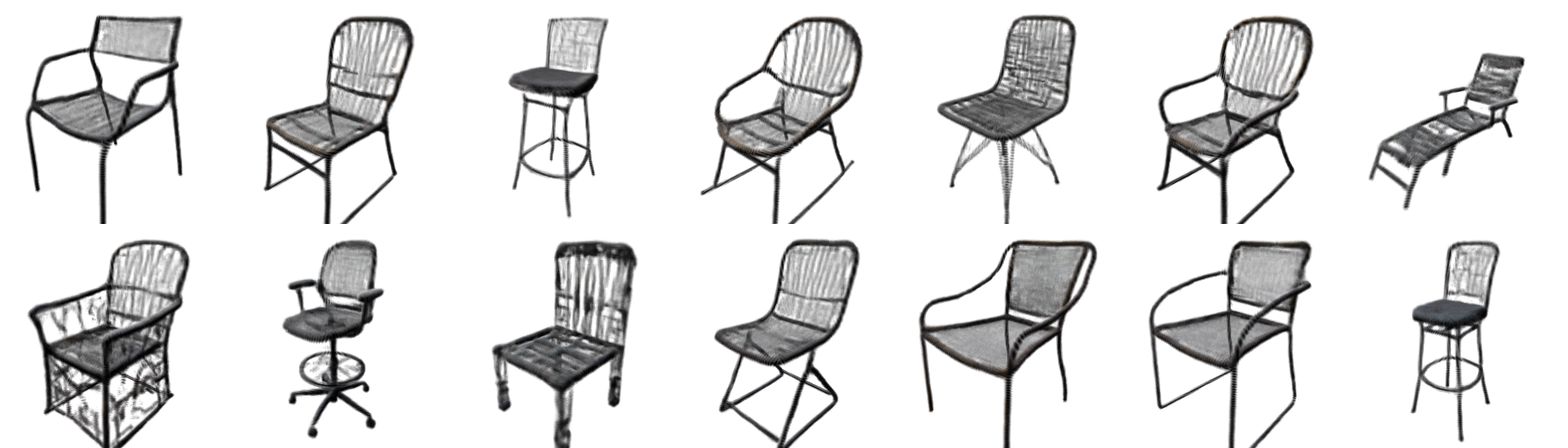}  \\ \midrule
              \raisebox{.5\height}{\includegraphics[width=0.06\linewidth]{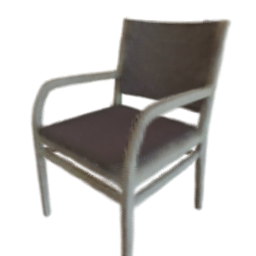}} &
              \includegraphics[width=0.4\linewidth]{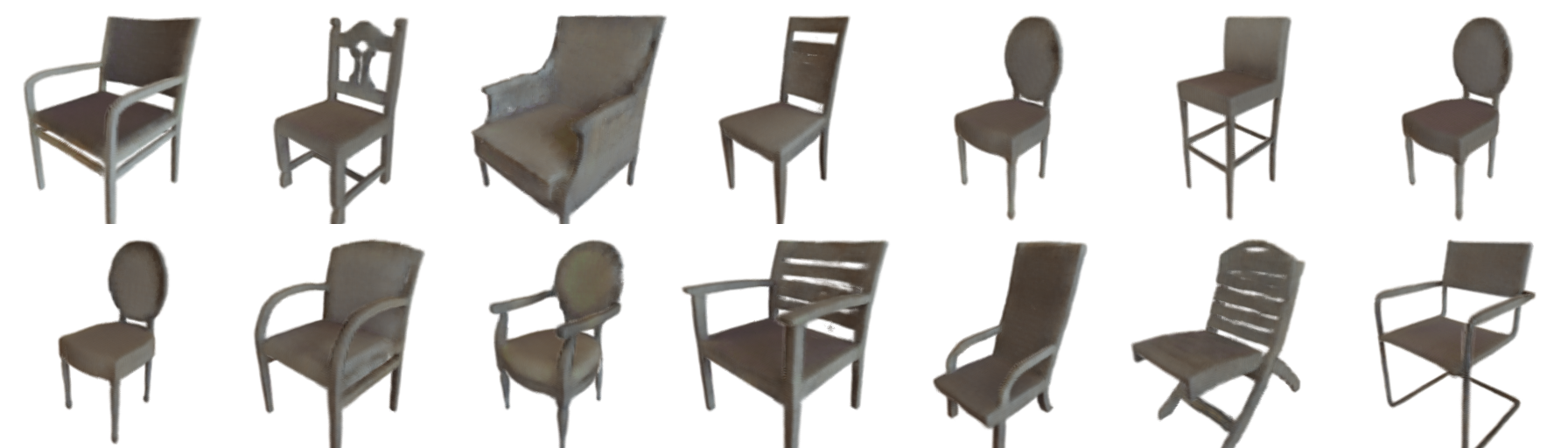} \\
              \raisebox{.5\height}{\includegraphics[width=0.06\linewidth]{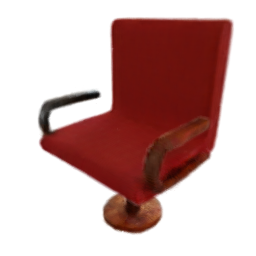}} &
              \includegraphics[width=0.4\linewidth]{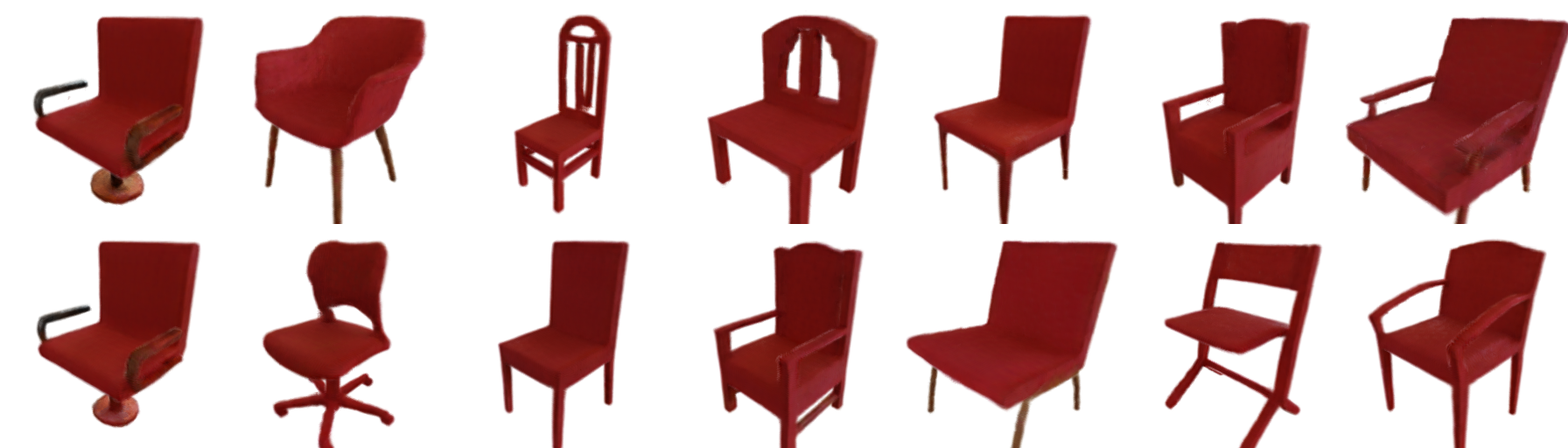}  \unskip
        \end{tabular}
        \label{subfig:DisentanglementShapeOnly}
        }
        \caption{\textbf{Qualitative examples of disentangled generation} on SRN cars, SRN chairs, PhotoShape chairs. \protect\subref{subfig:DisentanglementAppearanceOnly} \textbf{Appearance-only generation}: we show a generated object and objects with re-sampled appearance. \protect\subref{subfig:DisentanglementShapeOnly} \textbf{Shape-only generation}: we show a generated object and objects with re-sampled coarse shape.  We can get diverse samples of local appearance or coarse shape when the respective other is given.
        } 
        \label{fig:Disentanglement}
\end{figure*}

\paragraph{Training details.}
We construct neural point clouds by zero-initializing features for each point. In the Point-NeRF autodecoder, we optimize the reconstruction loss in \cref{eq:PointNeRFLoss} with the TV and KL regularizers from \cref{eq:PointNeRFTVLoss} and \cref{eq:PointNeRFKLLoss}. Further details on network architectures and training parameters are provided in the supplementals. For the diffusion model training, we normalize the neural point clouds and use DDPM~\cite{Ho2020} diffusion model parameters. Further details on the denoiser architecture, diffusion model parameters, and training parameters are provided in the supplementals. 

\subsection{Metrics}
\label{sec:metrics}

To measure the quality and diversity of the generated samples of the diffusion model, we report the FID~\cite{Heusel2017} and KID~\cite{Binkowski2018} metrics. FID and KID compare the appearance and diversity of two image sets by computing features for each image with an Inception model and comparing the feature distributions of the two sets. We use the images of the test set objects as the reference set. For comparability, we follow the evaluation procedures of previous works: on SRN Cars and Chairs, we generate the same number of objects as in the test set and render them from the same poses; on PhotoShape Chairs, we generate $1{,}000$ objects and render them from 10 poses that are sampled randomly from the Archimedean spiral poses. 
Furthermore, for the shape-only evaluation of our generated point clouds representing the coarse geometry, we employ 1-nearest-neighbor accuracy w.r.t. Chamfer and Earth Mover's Distance~\cite{Yang2019}. Last, we conduct a quantitative analysis by reporting the per-point mean cosine similarities between optimized neural point features of $10$ random training examples over $100$ different seeds with a fixed renderer. This measures the extent of \emph{many-to-one mappings}, i.e. how far features that represent the same appearance are away from each other. 
  
\subsection{Disentangled generation}
\label{sec:disentangled_generation}
A major advantage of using neural point clouds as 3D representation is their intrinsic disentanglement of shape and appearance: the point positions represent the coarse geometry and the features model local geometry and appearance. As described in \cref{sec:MethodDisentanledGeneration}, this property enables the proposed method to generate both modalities separately, even though a joint distribution is modeled with a single diffusion model. In the following, we present the results of this disentangled 3D shape and appearance generation. 

\methodname{} allows to explicitly control the point positions or point features throughout the diffusion process. As a consequence, we can re-sample appropriate features that fit to the given point positions or vice versa, resulting in generating new samples with one modality fixed. Results for the appearance-only generation are shown in \cref{subfig:DisentanglementAppearanceOnly} and for the shape-only generation in \cref{subfig:DisentanglementShapeOnly}. It can be seen that our method succeeds in generating diverse novel shapes or appearances when one modality is fixed. Note that our method performs an actual recombination and does more than retrieval of objects from the training dataset.

\begin{figure*}[!h]
      \centering
      \def\tmpwidth{0.28\linewidth}
    \begin{tabular}{@{}c|c@{}|c@{}}
      \raisebox{.8\height}{\rot{\small Cars}}\hspace{2mm}
      \includegraphics[width=\tmpwidth]{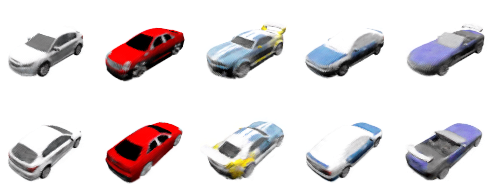} &
      \includegraphics[width=\tmpwidth]{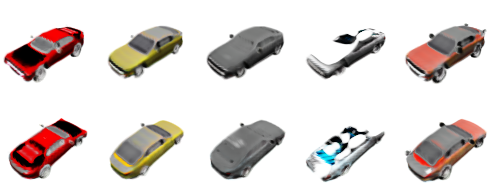} &
      \includegraphics[width=\tmpwidth]{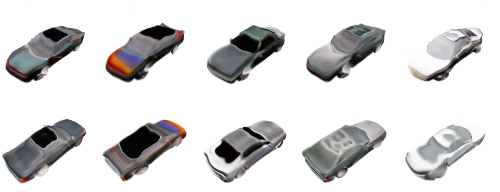} 
      \\
      \raisebox{.6\height}{\rot{\small Chairs}}\hspace{2mm}
      \includegraphics[width=\tmpwidth]{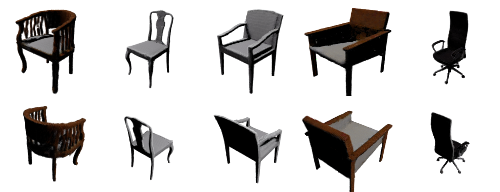} &
      \includegraphics[width=\tmpwidth]{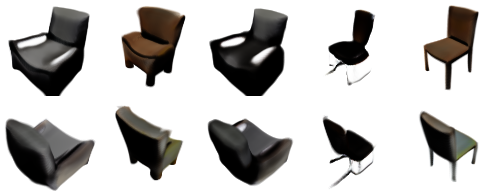} &
      \includegraphics[width=\tmpwidth]{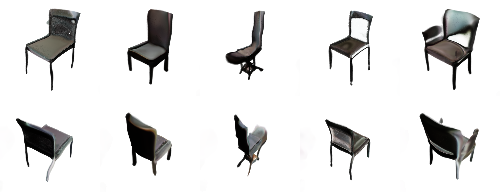} 
      \\
      \raisebox{.4\height}{\rot{\small PS Chairs}}\hspace{2mm}
      \includegraphics[width=\tmpwidth]{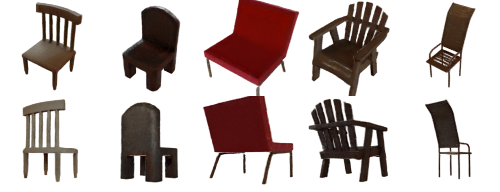} &
      \includegraphics[width=\tmpwidth]{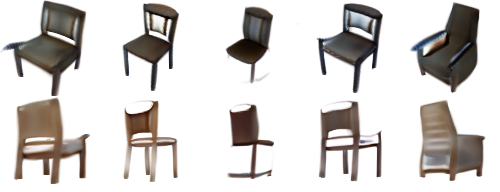} &
      \includegraphics[width=\tmpwidth]{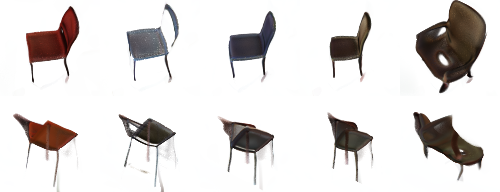} 
      \\
      Ours & GRAF~\cite{schwarz2020graf} & Disentangled3D~\cite{tewari2022disentangled3d}
  \end{tabular}
\caption{\textbf{Comparison against previous generative models that allow disentangled generation.}: While we present the first diffusion model allowing disentangled generation, earlier works are GAN-based. It can be seen that our model generates examples in much higher quality, as also evident from the metrics in Tab.~\ref{tab:DisnentangledComparison}. \vspace{-0.2cm}}
\label{fig:DisentanglementComparison}
\end{figure*}

Previous approaches that are able to generate disentangled 3D shape and appearance are GRAF~\cite{schwarz2020graf} and Disentangled3D (D3D)~\cite{tewari2022disentangled3d}, which are both GAN-based. We provide a quantitative comparison to these approaches regarding generation quality in Tab.~\ref{tab:DisnentangledComparison} and a qualitative comparison in Fig.~\ref{fig:DisentanglementComparison}. Our comparisons show that our proposed method is capable of disentangled generation with superior quality than these previous approaches.

\begin{table}[h!]
\small
\centering
\setlength{\tabcolsep}{0.7mm}
\scalebox{0.95}{
\begin{tabular}{l|
c c c c | c c
}

\toprule
    \textbf{Model}
    & \multicolumn{4}{c|}{\textbf{ShapeNet SRN}} 
    & \multicolumn{2}{c}{\textbf{PhotoShape}}
    \\
    
    & \multicolumn{2}{c}{\textbf{Cars}}
    & \multicolumn{2}{c|}{\textbf{Chairs}}
    & \multicolumn{2}{c}{\textbf{Chairs}}
    \\
    
    & FID$\downarrow$ & KID{\scriptsize/10\textsuperscript{\textminus3}}$\downarrow$
    & FID$\downarrow$ & KID{\scriptsize/10\textsuperscript{\textminus3}}$\downarrow$
    & FID$\downarrow$ & KID{\scriptsize/10\textsuperscript{\textminus3}}$\downarrow$
    \\
    
    \midrule

	GRAF~\cite{schwarz2020graf}
	& 40.95
	& 19.15
	& 37.19
	& 17.85
	& 34.49
	& 17.13
	\\

    D3D~\cite{tewari2022disentangled3d}
 	& 62.34
 	& 41.60
 	& 45.73
 	& 24.33
 	& 59.80
 	& 36.07
 	\\
    
    \midrule
    
\rowcolor{rowhighlight}	\methodname{} \scriptsize{(Ours)}
	& \textbf{28.38}
	& \textbf{17.62}
	& \textbf{9.87}
	& \textbf{3.62}
	& \textbf{14.45}
	& \textbf{5.40}
	\\
	
\bottomrule
\end{tabular}
}
\caption{\textbf{Comparison to disentanglement-capable approaches.} The numbers show that we clearly outperform previous generative models that allow disentangled generation.
\label{tab:DisnentangledComparison}
}
\end{table}

\subsection{3D diffusion comparison}
\label{sec:sota_comparison}
We compare \methodname{} with Functa~\cite{Dupont2022}, SSDNeRF~\cite{Chen2023}, and DiffRF~\cite{Muller2023}, previous works that generate 3D shape and appearance on medium-scale datasets with 3D diffusion models. 
We compare against SSDNeRF and Functa on SRN Cars and against DiffRF on Photoshape Chairs. 
Quantitative results are provided in \cref{tab:SOTAComparison}.
Our proposed method performs better than Functa and DiffRF methods and worse than SSDNeRF regarding the FID and KID metrics. However, none of these methods enable disentangled generation.
\begin{table}[h!]
\small
\centering
\setlength{\tabcolsep}{0.4mm}
\scalebox{0.97}{
\begin{tabular}{l|
c c
}

\toprule
    \textbf{Model}
    & \multicolumn{2}{c}{\textbf{PhotoShape Chairs}}
    \\

    & FID$\downarrow$ & KID{\scriptsize/10\textsuperscript{\textminus3}}$\downarrow$  
    \\
    
    \midrule

 	DiffRF
 	& 15.95
 	& 7.93
	\\
    
    \midrule
    
\rowcolor{rowhighlight}	 {\methodname{} \scriptsize{(Ours)}}
	& \textbf{14.45}
	& \textbf{5.40}
	\\
	
\bottomrule
\end{tabular}}
\hfill
\scalebox{0.97}{
\begin{tabular}{l|
c c
}

\toprule
    \textbf{Model}
    & \multicolumn{2}{c}{\textbf{SRN Cars}}
    \\

    & FID$\downarrow$ & KID{\scriptsize/10\textsuperscript{\textminus3}}$\downarrow$  
    \\
    
    \midrule

	Functa
	& 80.3
	& -
	\\

	SSDNeRF
	& \textbf{11.08}
	& \textbf{3.47}
	\\
    
    \midrule
    
\rowcolor{rowhighlight}	 {\methodname{} \scriptsize{(Ours)}}
	& 28.38
	& 17.62
	\\
	
\bottomrule
\end{tabular}
}

\caption{\textbf{Comparison to 3D diffusion models for unconditional 3D shape and appearance generation.} Our \methodname{} model achieves better scores than DiffRF and Functa. SSDNeRF performs slightly better. However, none of the other models enable disentangled generation.
\label{tab:SOTAComparison}
}
\end{table}

\subsection{Shape-only comparison}
\label{sec:shape_only_comparison}
Since our neural radiance field is build on top of a coarse point cloud, we evaluate the geometry of \methodname{} samples by comparing with the state of the art in point cloud generation in Tab.~\ref{tab:ShapeOnlyComparison}. Even though the point cloud defines only the coarse structure beneath a fine radiance field, the quality and diversity of our generated point clouds is comparable to the ones from approaches that are specialized for shape-only generation.
\begin{table}[!h]
\small
\centering
\setlength{\tabcolsep}{0.8mm}
\scalebox{0.97}{
\begin{tabular}{l|
c c
c c
}

\toprule
    \textbf{Model}
    & \multicolumn{2}{c}{\textbf{SRN Cars}}
    & \multicolumn{2}{c}{\textbf{SRN Chairs}}
    \\

    & CD$\downarrow$ & EMD$\downarrow$ 
    & CD$\downarrow$ & EMD$\downarrow$ 
    \\
    
    \midrule

	r-GAN~\cite{Achlioptas2018}
	& 94.46
	& 99.01
	& 83.69
 	& 99.70
	\\


	
	PointFlow~\cite{Yang2019}
	& 58.10
	& 56.25
	& 62.84
	& 60.57
	\\
	
	SoftFlow~\cite{Hyeongju2020}
	& 64.77
	& 60.09
	& 59.21
	& 60.05
	\\
	
	DPF-Net~\cite{Klokov2020}
	& 62.35
	& 54.48
	& 62.00
	& 58.53
	\\
	
	Shape-GF~\cite{Cai2020}
	& 63.20
	& 56.53
	& 68.96
	& 65.48
	\\
	
	PVD~\cite{Zhou2021}
	& \underline{54.55}
	& 53.83
	& \underline{56.26}
	& \underline{53.32}
	\\
    
    LION~\cite{Zeng2022}
    & \textbf{53.41}
    & \textbf{51.14}
    & \textbf{53.70}
    & \textbf{52.34}
    \\
    
    \midrule
    
\rowcolor{rowhighlight}	\methodname{} \scriptsize{(Ours)}
	& 60.23
	& \underline{52.41}
    & 60.50
    & 58.84
	\\
	
\bottomrule
\end{tabular}
}
\caption{
\textbf{Shape-only comparison.} We evaluate the point cloud generation part of our approach individually. Despite being just the coarse structure of a finer radiance field on top, \methodname{} can compete with the state of the art in point cloud generation.
\label{tab:ShapeOnlyComparison}
}
\end{table}

\subsection{Analysis}
\label{sec:Analyses}

As diffusion on hybrid point clouds and local radiance fields has not been done before, we conduct ablation studies and analyze various novel design choices. 
Here, we analyze the effects of different initialization strategies, feature dimensionality and regularization methods in the category-level Point-NeRF autodecoder and diffusion model. 
We provide a more detailed analysis in the supplementals. 

\paragraph{Neural point cloud initialization}
\label{sec:NeuralPointcloudInitialization}
Regarding the initialization of the neural point cloud features, we analyze initialization with features sampled from a Gaussian distribution against a zero initialization. Interestingly, we find that these different initializations strongly affect the feature space of the trained models, which directly translates to differences in generation quality (c.f. supplementals). \cref{tab:Cosine} indicates that the simple measure of zero initialization is able to largely mitigate the many-to-one mappings in the MLP decoder and provide much more coherent features. 

\paragraph{Neural point cloud regularization}
\label{sec:NeuralPointcloudRegularization}

As we assume that the structure of the feature space strongly affects the diffusion model, we analyze the effects of applying KL and TV regularizations to the neural point features during the Point-NeRF autodecoder training. 
\cref{tab:Cosine} shows that both regularizations further decrease the ambiguity of the latent space over the zero initialization. We can summarize that a combination of TV and KL regularization provides overall the best results (c.f. supplementals).
Overall, we found an appropriate initialization and regularization to be key ingredients for successful neural point cloud diffusion.

\begin{table}[!h]
\small
\centering
\begin{tabular}{
c
c
c|
c
c
c
}

\toprule
\textbf{Init.}
& \textbf{Reg.}
& $\lambda$
& \textbf{Cosine sim.}
\\

\midrule

Rand.
& \mn
& -
& 0.0306
\\

Zero
& \mn
& -
& 0.7695
\\

Zero
& TV
& 3.5e-6
& 0.9355
\\

Zero
& KL
& 1e-6
& 0.9480
\\

Zero
& TV,KL
& 3e-7,1e-7
& 0.9470
\\
	
\bottomrule
\end{tabular}
\caption{\textbf{Auto-decoded feature similarity.} We compute per-point mean cosine similarities between optimized neural point features of $10$ training examples for $100$ different seeds. Zero initialization and regularization effectively reduce the ambiguity of the auto-decoded latent space. This improves generation quality significantly, as shown in the supplementals.
\label{tab:Cosine}
}
\end{table}

\section{Conclusion}
\label{sec:Conclusion}
We presented a diffusion approach for neural point clouds with 
high-dimensional features that represent local parts of objects and can be rendered to images. We have shown that, in contrast to all other current 3D diffusion models, the neural point cloud enables our method to effectively \emph{disentangle} the coarse shape from the fine geometry and appearance and to control both factors separately. In experiments on ShapeNet and PhotoShape we could show that we clearly outperform previous methods that perform disentangled generation of 3D objects, while being competitive with unconditional generation methods. Further, we provide an extensive analysis and identified many-to-one mappings in auto-decoded latent spaces as the main challenge for the successful training of a latent diffusion model. Therefore, we proposed a suitable initialization and regularization of the neural point features as effective countermeasures. 

\paragraph{Acknowledgements.} We thank the authors of Disentangled3D for providing their code. The research leading to these results is funded by the Deutsche Forschungsgemeinschaft (DFG, German Research Foundation) under the project numbers 401269959 and 417962828. 
{
    \small
    \bibliographystyle{ieeenat_fullname}
    \bibliography{main}

\begin{thebibliography}{60}
\providecommand{\natexlab}[1]{#1}
\providecommand{\url}[1]{\texttt{#1}}
\expandafter\ifx\csname urlstyle\endcsname\relax
  \providecommand{\doi}[1]{doi: #1}\else
  \providecommand{\doi}{doi: \begingroup \urlstyle{rm}\Url}\fi

\bibitem[Achlioptas et~al.(2018)Achlioptas, Diamanti, Mitliagkas, and Guibas]{Achlioptas2018}
Panos Achlioptas, Olga Diamanti, Ioannis Mitliagkas, and Leonidas Guibas.
\newblock Learning representations and generative models for 3{D} point clouds.
\newblock In \emph{ICML}, 2018.

\bibitem[Anciukevi\v{c}ius et~al.(2023)Anciukevi\v{c}ius, Xu, Fisher, Henderson, Bilen, Mitra, and Guerrero]{Anciukevicius2023}
Titas Anciukevi\v{c}ius, Zexiang Xu, Matthew Fisher, Paul Henderson, Hakan Bilen, Niloy~J. Mitra, and Paul Guerrero.
\newblock Renderdiffusion: Image diffusion for 3d reconstruction, inpainting and generation.
\newblock In \emph{CVPR}, 2023.

\bibitem[Bińkowski et~al.(2018)Bińkowski, Sutherland, Arbel, and Gretton]{Binkowski2018}
Mikołaj Bińkowski, Danica~J. Sutherland, Michael Arbel, and Arthur Gretton.
\newblock Demystifying mmd gans.
\newblock In \emph{ICLR}, 2018.

\bibitem[Cai et~al.(2020)Cai, Yang, Averbuch-Elor, Hao, Belongie, Snavely, and Hariharan]{Cai2020}
Ruojin Cai, Guandao Yang, Hadar Averbuch-Elor, Zekun Hao, Serge Belongie, Noah Snavely, and Bharath Hariharan.
\newblock Learning gradient fields for shape generation.
\newblock In \emph{ECCV}, 2020.

\bibitem[Cao et~al.(2023)Cao, Kreis, Fidler, Sharp, and Yin]{cao2023texfusion}
Tianshi Cao, Karsten Kreis, Sanja Fidler, Nicholas Sharp, and KangXue Yin.
\newblock Texfusion: Synthesizing 3d textures with text-guided image diffusion models.
\newblock In \emph{ICCV}, 2023.

\bibitem[Chabra et~al.(2020)Chabra, Lenssen, Ilg, Schmidt, Straub, Lovegrove, and Newcombe]{deepls}
Rohan Chabra, Jan~E. Lenssen, Eddy Ilg, Tanner Schmidt, Julian Straub, Steven Lovegrove, and Richard Newcombe.
\newblock Deep local shapes: Learning local {SDF} priors for detailed {3D} reconstruction.
\newblock In \emph{ECCV}, 2020.

\bibitem[Chan et~al.(2022)Chan, Lin, Chan, Nagano, Pan, Mello, Gallo, Guibas, Tremblay, Khamis, Karras, and Wetzstein]{Chan2022}
Eric~R. Chan, Connor~Z. Lin, Matthew~A. Chan, Koki Nagano, Boxiao Pan, Shalini~De Mello, Orazio Gallo, Leonidas Guibas, Jonathan Tremblay, Sameh Khamis, Tero Karras, and Gordon Wetzstein.
\newblock Efficient geometry-aware {3D} generative adversarial networks.
\newblock In \emph{CVPR}, 2022.

\bibitem[Chang et~al.(2015)Chang, Funkhouser, Guibas, Hanrahan, Huang, Li, Savarese, Savva, Song, Su, Xiao, Yi, and Yu]{Chang2015}
Angel~X. Chang, Thomas Funkhouser, Leonidas Guibas, Pat Hanrahan, Qixing Huang, Zimo Li, Silvio Savarese, Manolis Savva, Shuran Song, Hao Su, Jianxiong Xiao, Li Yi, and Fisher Yu.
\newblock {ShapeNet}: An information-rich 3d model repository.
\newblock Technical Report \href{https://arxiv.org/abs/1512.03012}{arXiv:1512.03012}, Stanford University --- Princeton University --- Toyota Technological Institute at Chicago, 2015.

\bibitem[Chen et~al.(2023{\natexlab{a}})Chen, Siddiqui, Lee, Tulyakov, and Nie{\ss}ner]{chen2023text2tex}
Dave~Zhenyu Chen, Yawar Siddiqui, Hsin-Ying Lee, Sergey Tulyakov, and Matthias Nie{\ss}ner.
\newblock Text2tex: Text-driven texture synthesis via diffusion models.
\newblock In \emph{ICCV}, 2023{\natexlab{a}}.

\bibitem[Chen et~al.(2023{\natexlab{b}})Chen, Gu, Chen, Tian, Tu, Liu, and Su]{Chen2023}
Hansheng Chen, Jiatao Gu, Anpei Chen, Wei Tian, Zhuowen Tu, Lingjie Liu, and Hao Su.
\newblock Single-stage diffusion nerf: A unified approach to 3d generation and reconstruction.
\newblock In \emph{ICCV}, 2023{\natexlab{b}}.

\bibitem[Dao et~al.(2022)Dao, Fu, Ermon, Rudra, and R{\'e}]{dao2022}
Tri Dao, Daniel~Y. Fu, Stefano Ermon, Atri Rudra, and Christopher R{\'e}.
\newblock Flash{A}ttention: Fast and memory-efficient exact attention with {IO}-awareness.
\newblock In \emph{NeurIPS}, 2022.

\bibitem[Dhariwal and Nichol(2021)]{diffbeatgan}
Prafulla Dhariwal and Alexander Nichol.
\newblock Diffusion models beat gans on image synthesis.
\newblock In \emph{NeurIPS}, 2021.

\bibitem[Dupont et~al.(2022)Dupont, Kim, Eslami, Rezende, and Rosenbaum]{Dupont2022}
Emilien Dupont, Hyunjik Kim, S.~M.~Ali Eslami, Danilo~Jimenez Rezende, and Dan Rosenbaum.
\newblock From data to functa: Your data point is a function and you can treat it like one.
\newblock In \emph{ICML}, 2022.

\bibitem[Erkoç et~al.(2023)Erkoç, Ma, Shan, Nießner, and Dai]{Erkoc2023}
Ziya Erkoç, Fangchang Ma, Qi Shan, Matthias Nießner, and Angela Dai.
\newblock {HyperDiffusion}: Generating implicit neural fields with weight-space diffusion, 2023.

\bibitem[Heusel et~al.(2017)Heusel, Ramsauer, Unterthiner, Nessler, and Hochreiter]{Heusel2017}
Martin Heusel, Hubert Ramsauer, Thomas Unterthiner, Bernhard Nessler, and Sepp Hochreiter.
\newblock Gans trained by a two time-scale update rule converge to a local nash equilibrium.
\newblock In \emph{NeurIPS}, 2017.

\bibitem[Ho et~al.(2020)Ho, Jain, and Abbeel]{Ho2020}
Jonathan Ho, Ajay Jain, and Pieter Abbeel.
\newblock Denoising diffusion probabilistic models.
\newblock In \emph{NeurIPS}, 2020.

\bibitem[Jang and Agapito(2021)]{jang2021codenerf}
Wonbong Jang and Lourdes Agapito.
\newblock Codenerf: Disentangled neural radiance fields for object categories.
\newblock In \emph{ICCV}, 2021.

\bibitem[Jun and Nichol(2023)]{Jun2023}
Heewoo Jun and Alex Nichol.
\newblock {Shap-E}: Generating conditional 3d implicit functions.
\newblock \emph{\href{https://arxiv.org/abs/2305.02463}{arXiv:2305.02463}}, 2023.

\bibitem[Karras et~al.(2022)Karras, Aittala, Aila, and Laine]{Karras2022}
Tero Karras, Miika Aittala, Timo Aila, and Samuli Laine.
\newblock Elucidating the design space of diffusion-based generative models.
\newblock In \emph{NeurIPS}, 2022.

\bibitem[Kim et~al.(2020)Kim, Lee, Kang, Lee, and Kim]{Hyeongju2020}
Hyeongju Kim, Hyeonseung Lee, Woo~Hyun Kang, Joun~Yeop Lee, and Nam~Soo Kim.
\newblock Softflow: Probabilistic framework for normalizing flow on manifolds.
\newblock In \emph{NeurIPS}, 2020.

\bibitem[Kingma and Welling(2014)]{Kingma2014}
Diederik~P. Kingma and Max Welling.
\newblock {Auto-Encoding Variational Bayes}.
\newblock In \emph{ICLR}, 2014.

\bibitem[Klokov et~al.(2020)Klokov, Boyer, and Verbeek]{Klokov2020}
Roman Klokov, Edmond Boyer, and Jakob Verbeek.
\newblock Discrete point flow networks for efficient point cloud generation.
\newblock In \emph{ECCV}, 2020.

\bibitem[Li et~al.(2023)Li, Duan, Zhou, and Lu]{li2023diffusionsdf}
Muheng Li, Yueqi Duan, Jie Zhou, and Jiwen Lu.
\newblock Diffusion-sdf: Text-to-shape via voxelized diffusion.
\newblock In \emph{CVPR}, 2023.

\bibitem[Lin et~al.(2023)Lin, Gao, Tang, Takikawa, Zeng, Huang, Kreis, Fidler, Liu, and Lin]{Lin2023}
Chen-Hsuan Lin, Jun Gao, Luming Tang, Towaki Takikawa, Xiaohui Zeng, Xun Huang, Karsten Kreis, Sanja Fidler, Ming-Yu Liu, and Tsung-Yi Lin.
\newblock Magic3d: High-resolution text-to-3d content creation.
\newblock In \emph{CVPR}, 2023.

\bibitem[Liu et~al.(2023)Liu, Wu, Van~Hoorick, Tokmakov, Zakharov, and Vondrick]{Liu2023}
Ruoshi Liu, Rundi Wu, Basile Van~Hoorick, Pavel Tokmakov, Sergey Zakharov, and Carl Vondrick.
\newblock Zero-1-to-3: Zero-shot one image to 3d object.
\newblock In \emph{ICCV}, 2023.

\bibitem[Lugmayr et~al.(2022)Lugmayr, Danelljan, Romero, Yu, Timofte, and Van~Gool]{Lugmayr2022}
Andreas Lugmayr, Martin Danelljan, Andres Romero, Fisher Yu, Radu Timofte, and Luc Van~Gool.
\newblock {RePaint}: Inpainting using denoising diffusion probabilistic models.
\newblock In \emph{CVPR}, 2022.

\bibitem[Luo and Hu(2021)]{Luo2021}
Shitong Luo and Wei Hu.
\newblock Diffusion probabilistic models for 3d point cloud generation.
\newblock In \emph{CVPR}, 2021.

\bibitem[Melas-Kyriazi et~al.(2023)Melas-Kyriazi, Laina, Rupprecht, and Vedaldi]{Melas2023}
Luke Melas-Kyriazi, Iro Laina, Christian Rupprecht, and Andrea Vedaldi.
\newblock Realfusion: 360deg reconstruction of any object from a single image.
\newblock In \emph{CVPR}, 2023.

\bibitem[Mildenhall et~al.(2020)Mildenhall, Srinivasan, Tancik, Barron, Ramamoorthi, and Ng]{Mildenhall2020}
Ben Mildenhall, Pratul~P. Srinivasan, Matthew Tancik, Jonathan~T. Barron, Ravi Ramamoorthi, and Ren Ng.
\newblock Nerf: Representing scenes as neural radiance fields for view synthesis.
\newblock In \emph{ECCV}, 2020.

\bibitem[M{\"u}ller et~al.(2023)M{\"u}ller, Siddiqui, Porzi, Bulo, Kontschieder, and Nie{\ss}ner]{Muller2023}
Norman M{\"u}ller, Yawar Siddiqui, Lorenzo Porzi, Samuel~Rota Bulo, Peter Kontschieder, and Matthias Nie{\ss}ner.
\newblock Diffrf: Rendering-guided 3d radiance field diffusion.
\newblock In \emph{CVPR}, 2023.

\bibitem[Nichol et~al.(2022)Nichol, Jun, Dhariwal, Mishkin, and Chen]{Nichol2022}
Alex Nichol, Heewoo Jun, Prafulla Dhariwal, Pamela Mishkin, and Mark Chen.
\newblock {Point-E}: A system for generating 3d point clouds from complex prompts.
\newblock \emph{\href{https://arxiv.org/abs/2212.08751}{arXiv:2212.08751}}, 2022.

\bibitem[Niemeyer and Geiger(2021)]{niemeyer2021giraffe}
Michael Niemeyer and Andreas Geiger.
\newblock Giraffe: Representing scenes as compositional generative neural feature fields.
\newblock In \emph{CVPR}, 2021.

\bibitem[Niemeyer et~al.(2020)Niemeyer, Mescheder, Oechsle, and Geiger]{niemeyer2020differentiable}
Michael Niemeyer, Lars Mescheder, Michael Oechsle, and Andreas Geiger.
\newblock Differentiable volumetric rendering: Learning implicit 3d representations without 3d supervision.
\newblock In \emph{CVPR}, 2020.

\bibitem[Park et~al.(2018)Park, Rematas, Farhadi, and Seitz]{Park2018}
Keunhong Park, Konstantinos Rematas, Ali Farhadi, and Steven~M. Seitz.
\newblock Photoshape: Photorealistic materials for large-scale shape collections.
\newblock \emph{ACM TOG}, 2018.

\bibitem[Peebles and Xie(2023)]{Peebles2022DiT}
William Peebles and Saining Xie.
\newblock Scalable diffusion models with transformers.
\newblock In \emph{ICCV}, 2023.

\bibitem[Poole et~al.(2023)Poole, Jain, Barron, and Mildenhall]{Poole2023}
Ben Poole, Ajay Jain, Jonathan~T. Barron, and Ben Mildenhall.
\newblock Dreamfusion: Text-to-3d using 2d diffusion.
\newblock In \emph{ICLR}, 2023.

\bibitem[Ramesh et~al.(2021)Ramesh, Pavlov, Goh, Gray, Voss, Radford, Chen, and Sutskever]{dalle}
Aditya Ramesh, Mikhail Pavlov, Gabriel Goh, Scott Gray, Chelsea Voss, Alec Radford, Mark Chen, and Ilya Sutskever.
\newblock Zero-shot text-to-image generation.
\newblock In \emph{ICML}, 2021.

\bibitem[Rombach et~al.(2022)Rombach, Blattmann, Lorenz, Esser, and Ommer]{Rombach2022}
Robin Rombach, Andreas Blattmann, Dominik Lorenz, Patrick Esser, and Björn Ommer.
\newblock High-resolution image synthesis with latent diffusion models.
\newblock In \emph{CVPR}, 2022.

\bibitem[Saharia et~al.(2022)Saharia, Chan, Saxena, Li, Whang, Denton, Ghasemipour, Gontijo~Lopes, Karagol~Ayan, Salimans, et~al.]{imagegen}
Chitwan Saharia, William Chan, Saurabh Saxena, Lala Li, Jay Whang, Emily~L Denton, Kamyar Ghasemipour, Raphael Gontijo~Lopes, Burcu Karagol~Ayan, Tim Salimans, et~al.
\newblock Photorealistic text-to-image diffusion models with deep language understanding.
\newblock In \emph{NeurIPS}, 2022.

\bibitem[Schwarz et~al.(2020)Schwarz, Liao, Niemeyer, and Geiger]{schwarz2020graf}
Katja Schwarz, Yiyi Liao, Michael Niemeyer, and Andreas Geiger.
\newblock Graf: Generative radiance fields for 3d-aware image synthesis.
\newblock In \emph{NeurIPS}, 2020.

\bibitem[Shim et~al.(2023)Shim, Kang, and Joo]{shim2023diffusion}
Jaehyeok Shim, Changwoo Kang, and Kyungdon Joo.
\newblock Diffusion-based signed distance fields for 3d shape generation.
\newblock In \emph{CVPR}, 2023.

\bibitem[Shue et~al.(2023{\natexlab{a}})Shue, Chan, Po, Ankner, Wu, and Wetzstein]{Shue2023}
J.~Ryan Shue, Eric~Ryan Chan, Ryan Po, Zachary Ankner, Jiajun Wu, and Gordon Wetzstein.
\newblock 3d neural field generation using triplane diffusion.
\newblock In \emph{CVPR}, 2023{\natexlab{a}}.

\bibitem[Shue et~al.(2023{\natexlab{b}})Shue, Chan, Po, Ankner, Wu, and Wetzstein]{Shue_2023_CVPR}
J.~Ryan Shue, Eric~Ryan Chan, Ryan Po, Zachary Ankner, Jiajun Wu, and Gordon Wetzstein.
\newblock 3d neural field generation using triplane diffusion.
\newblock In \emph{CVPR}, 2023{\natexlab{b}}.

\bibitem[Sitzmann et~al.(2019)Sitzmann, Zollh{\"o}fer, and Wetzstein]{Sitzmann2019}
Vincent Sitzmann, Michael Zollh{\"o}fer, and Gordon Wetzstein.
\newblock Scene representation networks: Continuous 3d-structure-aware neural scene representations.
\newblock In \emph{NeurIPS}, 2019.

\bibitem[Sohl-Dickstein et~al.(2015)Sohl-Dickstein, Weiss, Maheswaranathan, and Ganguli]{Sohldickstein2015}
Jascha Sohl-Dickstein, Eric Weiss, Niru Maheswaranathan, and Surya Ganguli.
\newblock Deep unsupervised learning using nonequilibrium thermodynamics.
\newblock In \emph{ICML}, 2015.

\bibitem[Song and Ermon(2019)]{Song2019}
Yang Song and Stefano Ermon.
\newblock Generative modeling by estimating gradients of the data distribution.
\newblock In \emph{NeurIPS}, 2019.

\bibitem[Tewari et~al.(2022)Tewari, {B R}, Pan, Fried, Agrawala, and Theobalt]{tewari2022disentangled3d}
Ayush Tewari, MalliKarjun {B R}, Xingang Pan, Ohad Fried, Maneesh Agrawala, and Christian Theobalt.
\newblock Disentangled3d: Learning a 3d generative model with disentangled geometry and appearance from monocular images.
\newblock In \emph{CVPR}, 2022.

\bibitem[Vaswani et~al.(2017)Vaswani, Shazeer, Parmar, Uszkoreit, Jones, Gomez, Kaiser, and Polosukhin]{Vaswani2017}
Ashish Vaswani, Noam Shazeer, Niki Parmar, Jakob Uszkoreit, Llion Jones, Aidan~N Gomez, \L~ukasz Kaiser, and Illia Polosukhin.
\newblock Attention is all you need.
\newblock In \emph{NeurIPS}, 2017.

\bibitem[Wewer et~al.(2023)Wewer, Ilg, Schiele, and Lenssen]{wewer23simnp}
Christopher Wewer, Eddy Ilg, Bernt Schiele, and Jan~Eric Lenssen.
\newblock Simnp: Learning self-similarity priors between neural points.
\newblock In \emph{ICCV}, 2023.

\bibitem[Xu et~al.(2022)Xu, Xu, Philip, Bi, Shu, Sunkavalli, and Neumann]{Xu2022}
Qiangeng Xu, Zexiang Xu, Julien Philip, Sai Bi, Zhixin Shu, Kalyan Sunkavalli, and Ulrich Neumann.
\newblock {Point-NeRF}: Point-based neural radiance fields.
\newblock In \emph{CVPR}, 2022.

\bibitem[Yang et~al.(2019)Yang, Huang, Hao, Liu, Belongie, and Hariharan]{Yang2019}
Guandao Yang, Xun Huang, Zekun Hao, Ming-Yu Liu, Serge Belongie, and Bharath Hariharan.
\newblock Pointflow: 3d point cloud generation with continuous normalizing flows.
\newblock In \emph{ICCV}, 2019.

\bibitem[Yao et~al.(2018)Yao, Luo, Li, Fang, and Quan]{mvsnet}
Yao Yao, Zixin Luo, Shiwei Li, Tian Fang, and Long Quan.
\newblock Mvsnet: Depth inference for unstructured multi-view stereo.
\newblock In \emph{ECCV}, 2018.

\bibitem[Youwang et~al.(2024)Youwang, Oh, and Pons-Moll]{youwang2023paintit}
Kim Youwang, Tae-Hyun Oh, and Gerard Pons-Moll.
\newblock Paint-it: Text-to-texture synthesis via deep convolutional texture map optimization and physically-based rendering.
\newblock In \emph{CVPR}, 2024.

\bibitem[Zeng et~al.(2022)Zeng, Vahdat, Williams, Gojcic, Litany, Fidler, and Kreis]{Zeng2022}
Xiaohui Zeng, Arash Vahdat, Francis Williams, Zan Gojcic, Or Litany, Sanja Fidler, and Karsten Kreis.
\newblock {LION}: Latent point diffusion models for 3d shape generation.
\newblock In \emph{NeurIPS}, 2022.

\bibitem[Zhang et~al.(2023)Zhang, Tang, Niessner, and Wonka]{zhang20233dshape2vecset}
Biao Zhang, Jiapeng Tang, Matthias Niessner, and Peter Wonka.
\newblock 3dshape2vecset: A 3d shape representation for neural fields and generative diffusion models.
\newblock \emph{ACM TOG}, 2023.

\bibitem[Zhang and Agrawala(2023)]{controlnet}
Lvmin Zhang and Maneesh Agrawala.
\newblock Adding conditional control to text-to-image diffusion models.
\newblock \emph{\href{https://arxiv.org/abs/2302.05543}{arXiv preprint arXiv:2302.05543}}, 2023.

\bibitem[Zheng et~al.(2023)Zheng, Pan, Wang, Tong, Liu, and Shum]{zheng2023lasdiffusion}
Xin-Yang Zheng, Hao Pan, Peng-Shuai Wang, Xin Tong, Yang Liu, and Heung-Yeung Shum.
\newblock Locally attentional sdf diffusion for controllable 3d shape generation.
\newblock \emph{ACM TOG}, 2023.

\bibitem[Zhou et~al.(2021)Zhou, Du, and Wu]{Zhou2021}
Linqi Zhou, Yilun Du, and Jiajun Wu.
\newblock 3d shape generation and completion through point-voxel diffusion.
\newblock In \emph{ICCV}, 2021.

\bibitem[Zhou and Tulsiani(2023)]{Zhou2023}
Zhizhuo Zhou and Shubham Tulsiani.
\newblock Sparsefusion: Distilling view-conditioned diffusion for 3d reconstruction.
\newblock In \emph{CVPR}, 2023.

\bibitem[Zhu et~al.(2018)Zhu, Zhang, Zhang, Wu, Torralba, Tenenbaum, and Freeman]{zhu2018visual}
Jun-Yan Zhu, Zhoutong Zhang, Chengkai Zhang, Jiajun Wu, Antonio Torralba, Josh Tenenbaum, and Bill Freeman.
\newblock Visual object networks: Image generation with disentangled 3d representations.
\newblock In \emph{NeurIPS}, 2018.

\end{thebibliography}
}

\clearpage
\renewcommand{\thesection}{A\arabic{section}} \setcounter{section}{0} \renewcommand{\thefigure}{A\arabic{figure}} \setcounter{figure}{0} \renewcommand{\thetable}{A\arabic{table}} \setcounter{table}{0}
\setcounter{page}{1}
\onecolumn
{  
\centering
\Large
\textbf{\thetitle}\\
\vspace{0.5em}Supplementary Material \\
\vspace{1.0em}
}

\section{Visualization of the neural point cloud diffusion process}

\begin{figure*}[h!]
        \centering
        \includegraphics[width=0.99\linewidth]{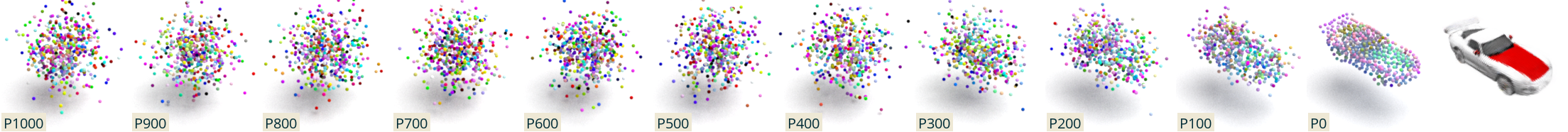}\\
        \includegraphics[width=0.99\linewidth]{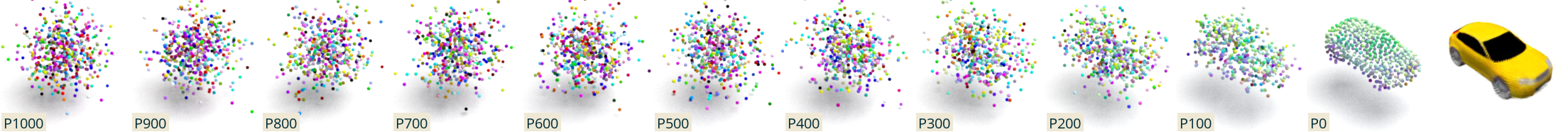}\\ \hspace{1cm}
        \includegraphics[width=0.99\linewidth]{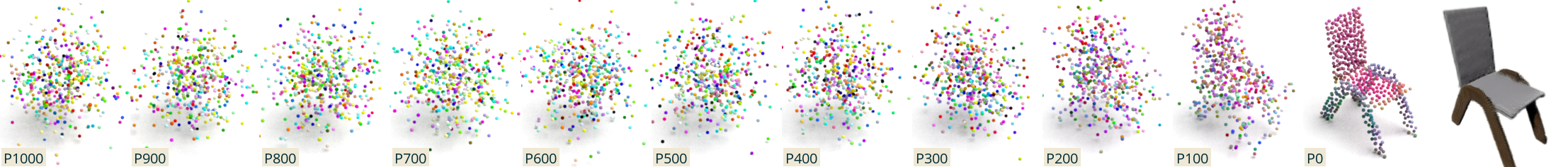}\\
        \includegraphics[width=0.99\linewidth]{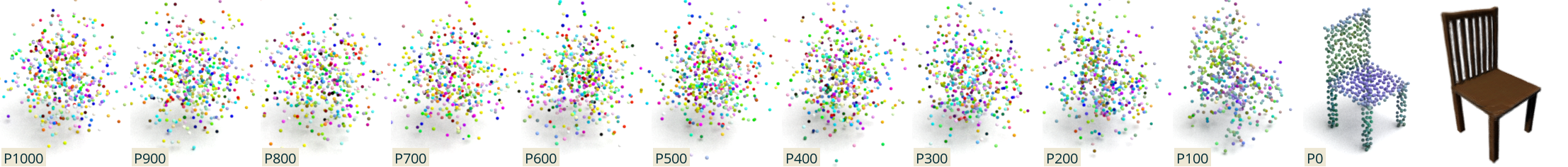}\\ \hspace{1cm}
        \includegraphics[width=0.99\linewidth]{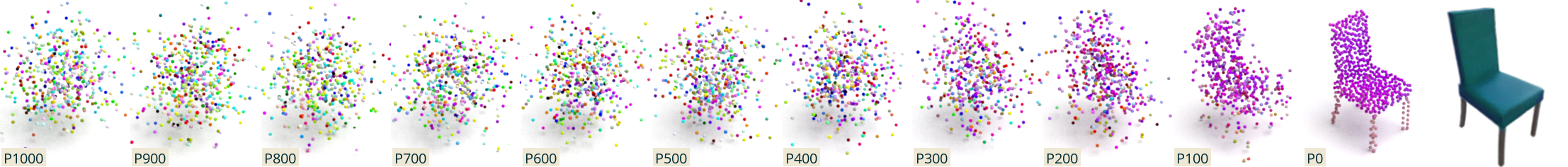}\\
        \includegraphics[width=0.99\linewidth]{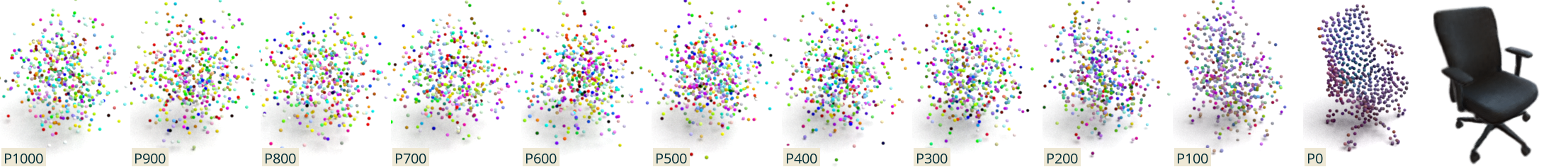}
        \caption{\textbf{Visualization of the neural point cloud diffusion process}. We generate the shape and appearance of 3D objects on ShapeNet Cars, ShapeNet Chairs, and PhotoShape Chairs with the proposed Neural Point Cloud Diffusion (\methodname{}) model. We visualize the neural point clouds $\neuralpointcloud_t = (\posmatrix_t, \fmatrix_t)$ from intermediate timesteps $t$ of the diffusion process. In total, the diffusion process of \methodname{} has 1000 timesteps and we visualize every 100th timestep. The features of the neural point clouds are visualized by taking the first three PCA components as RGB color. The last visualized neural point cloud $\neuralpointcloud_0$ represents the final generated 3D object. Additionally, we visualize a Point-NeRF rendering of the final neural point cloud.}
        \label{fig:NeuralPointCloudDiffusion}
\end{figure*}

Figure~\ref{fig:NeuralPointCloudDiffusion} shows a visualization of the neural point cloud diffusion process for unconditional generation on ShapeNet Cars, ShapeNet Chairs, and PhotoShape Chairs.

\newpage

\section{Implementation details}

\subsection{Category-Level Point-NeRF Autodecoder}

\paragraph{Architecture.} For the aggregation MLP $\localmlp$, we use 4 linear layers with a hidden dimension of 256, each followed by a LeakyReLU, and an output projection linear layer that maps to 256d. For the color MLP $\colormlp$, we use 4 linear layers with a hidden dimension of 256, each followed by a LeakyReLU, and an output projection linear layer that maps to 3d. For the density MLP $\shapemlp$, we use 1 linear layer with a hidden dimension of 256, followed by a LeakyReLU, and an output projection linear layer that maps to 1d.

\paragraph{Training parameters.} We construct training samples by splitting the available views per object into groups of 50 views per sample. In each iteration, we use 8 samples and 112 pixels of each view in each sample. The image reconstruction loss is hence optimized for an effective batch size of $8\cdot 50\cdot 112=44800$ pixels. For the volumetric rendering of each ray, we sample 128 shading points. We use the Adam optimizer with a constant learning rate of $1e\!\!-\!\!3$. On SRN cars and chairs, we train Point-NeRF for ca. 7500 epochs. On PhotoShape Chairs, we train Point-NeRF for ca. 1875 epochs (due to the 4 times higher number of training views per object, each epoch contains 4 times as many samples per object as in SRN Cars/Chairs). The training time on a single RTX 4090 GPU is between 3 and 5 days, depending on the dataset. 

\paragraph{Rendering runtime.} For the Point-NeRF rendering at a resolution of 128x128px, we measure a mean runtime of 35 msec on a RTX 4090 GPU.

\subsection{Diffusion model}

\paragraph{Architecture.} For the denoiser network of the diffusion model, we use a standard transformer architecture with 24 layers, a feature dimension of 1024d and 16 heads~\cite{Nichol2022, Peebles2022DiT}. This architecture has ca. 300M parameters.

\paragraph{Diffusion model parameters.} For the diffusion model, we use the linear noise schedule from DDPM~\cite{Ho2020} with 1,000 steps and $\beta$ ranging from $0.0001$ to $0.02$. We normalize the neural point clouds such that the positions are unit Gaussian distributed and the features are in the range $[-1, 1]$ (and apply the inverse transform later before rendering the representation). During sampling, we clip the coordinates and features to the respective minimum and maximum values of the training dataset. 

\paragraph{Training parameters.} We train the diffusion model for ca 1.8M iterations with a batch size of 32. We train on a single RTX 4090 GPU with 16 bit and employ flash attention~\cite{dao2022}. Training takes ca. 8 days. We use an exponential moving average over the model parameters with a decay of of 0.9999. On ShapeNet Cars, we use a constant learning rate of $7e\!\!-\!\!5$. On ShapeNet Chairs and PhotoShape Chairs, we use a lower constant learning rate of $4e\!\!-\!\!5$, as we observed instablities during training. On all datasets, we use a weight decay of 0.01. 

\paragraph{Runtime for unconditional generation.} For unconditional generation with 1000 diffusion steps, we measure a mean runtime of 8.56 sec for a single generation (batch size 1) on a RTX 4090 GPU.

\newcommand{\NumFreeCoordSteps}{N_{\textrm{rev}}}
\newcommand{\NumRepaintStepsFeats}{N_{\textrm{repaint}}}
\newcommand{\NumRepaintsFeats}{N_{\textrm{resample}}}
\subsection{Disentangled generation} 
As described in the main paper, our disentangled generation is comparable to masked image inpainting, using an approach similar to RePaint~\cite{Lugmayr2022}. Instead of masking image parts, we mask one modality of our representation (point positions or features). In \cref{alg:disentangled_generation}, we provide the algorithm that we use for disentangled generation in full detail. The algorithm is for appearance-only generation, \ie where point positions $\posmatrix_0$ are given and the goal is to generate point features $\fmatrix_0$. For shape-only generation, it works vice-versa.

\paragraph{Initialization.} As described in the main paper, we obtain the initial noisy neural point cloud $(\posmatrix_T, \fmatrix_T)$ by sampling $\fmatrix_T$ from a unit Gaussian distribution and computing $\posmatrix_T$ from $\posmatrix_0$ via the forward diffusion process. In \cref{alg:disentangled_generation}, this is described by line 2. Importantly, for computing point positions via the forward diffusion process, we sample noise $\mathbf{\epsilon}^{\posmatrix} \sim \mathcal{N}(\mathbf{0}, \mathbf{I})$ once in the beginning and re-use it in subsequent diffusion steps.

\paragraph{Diffusion process.} Throughout the diffusion process, we update the point features $\fmatrix_{t-1}$ from the denoiser outputs according to the reverse diffusion process (lines 4 and 5). 

We update the point positions $\posmatrix_{t-1}$ from the given $\posmatrix_0$ via the forward diffusion process in instead of the denoiser outputs (line 10). For this, we re-use the noise $\mathbf{\epsilon}^{\posmatrix}$ that we sampled during the initialization.

\paragraph{Reverse process also for the point positions.} In the last $\NumFreeCoordSteps$ steps of the diffusion process, we also use the outputs of the denoiser network to update the point positions via the reverse process (line 13). Depending on the chosen $\NumFreeCoordSteps$ this allows for a trade-off between better coherence in the generations at the cost of deviations from the given input point positions. 

\paragraph{Resampling.} Further, as in RePaint, we apply $\NumRepaintsFeats$ resampling steps to the denoised point features in the last $\NumRepaintStepsFeats$ steps of the diffusion process (lines 15 to 21). We observe that this is helpful for the coherence of the generations (in accordance with Fig.~3 of the RePaint paper). As we use the reverse process for the point positions in the last  $\NumFreeCoordSteps$ steps of the diffusion process, we do not use resampling in these iterations (second condition in line 15). 

\begin{algorithm}[!h]
\footnotesize
\caption{Appearance-only generation} \label{alg:disentangled_generation}
\begin{algorithmic}[1]

\State \textbf{Input:} Point Positions $\posmatrix_0$, $\NumFreeCoordSteps$, $\NumRepaintStepsFeats$, $\NumRepaintsFeats$
\Statex

\State \textbf{Initialization:} $ \mathbf{\epsilon}^{\posmatrix} \sim \mathcal{N}(\mathbf{0}, \mathbf{I}) \quad \posmatrix_T = \sqrt{\bar\alpha_T} \posmatrix_0 + \sqrt{1-\bar\alpha_T}\mathbf{\epsilon}^{\posmatrix} \quad \fmatrix_T \sim \mathcal{N}(\mathbf{0}, \mathbf{I})$
\Statex

\For{$t=T, \dotsc, 1$}\Comment{we use $T=1000$}

\State $(\mathbf{\epsilon}^{\posmatrix}_\theta, \mathbf{\epsilon}^{\fmatrix}_\theta)=T_\theta((\posmatrix_t, \fmatrix_t), t)$ \Comment{estimate noise with denoiser $T_\theta$}
\Statex

\State $\fmatrix_{t-1} {=} \frac{1}{\sqrt{\alpha_t}} \left(\fmatrix_{t} {-} \frac{\beta_t}{\sqrt{1-\bar{\alpha_t}}}\mathbf{\epsilon}^{\fmatrix}_\theta\right){+}\frac{1-\bar{\alpha}_{t-1}}{1-\bar{\alpha}_t}\beta_t\mathbf{\epsilon} \quad \mathbf{\epsilon} {\sim} \mathcal{N}(\mathbf{0}, \mathbf{I})$ 
\Statex

\If{$t > \NumFreeCoordSteps$} \Comment{compute $\posmatrix_{t-1}$ via forward process}
\If{$t == 1$}
\State $ \posmatrix_{t-1} = \posmatrix_{0}$
\Else
\State $ \posmatrix_{t-1} = \sqrt{\bar\alpha_{t-1}} \posmatrix_0 + \sqrt{1-\bar\alpha_{t-1}}\mathbf{\epsilon}^{\posmatrix}$ 
\EndIf
\Else \Comment{compute $\posmatrix_{t-1}$ via reverse process for last $\NumFreeCoordSteps$ steps}
\State $\posmatrix_{t-1} {=} \frac{1}{\sqrt{\alpha_t}} \left(\posmatrix_{t} {-} \frac{\beta_t}{\sqrt{1-\bar{\alpha_t}}}\mathbf{\epsilon}^{\posmatrix}_\theta\right){+}\frac{1-\bar{\alpha}_{t-1}}{1-\bar{\alpha}_t}\beta_t\mathbf{\epsilon} \quad \mathbf{\epsilon} {\sim} \mathcal{N}(\mathbf{0}, \mathbf{I})$ 
\EndIf
\Statex

\If{$t \leq \NumRepaintStepsFeats$ and $t \geq \NumFreeCoordSteps$} \Comment{use RePaint-resampling in last $\NumRepaintStepsFeats$ steps in case $\posmatrix_t$ comes from forward process}
\For{$1, \dotsc, \NumRepaintsFeats$}\Comment{resample $\NumRepaintsFeats$ times}
\State $\fmatrix_{t} \sim \mathcal{N}(\fmatrix_{t}; \sqrt{1-\beta_t} \fmatrix_{t-1}; \beta_t\mathbf{I})$ \Comment{forward process}
\State $(\mathbf{\epsilon}^{\posmatrix}_\theta, \mathbf{\epsilon}^{\fmatrix}_\theta)=T_\theta((\posmatrix_t, \fmatrix_t), t)$ \Comment{denoise again}
\State $\fmatrix_{t-1} {=} \frac{1}{\sqrt{\alpha_t}} \left(\fmatrix_{t} {-} \frac{\beta_t}{\sqrt{1-\bar{\alpha_t}}}\mathbf{\epsilon}^{\fmatrix}_\theta\right){+}\frac{1-\bar{\alpha}_{t-1}}{1-\bar{\alpha}_t}\beta_t\mathbf{\epsilon} \quad \mathbf{\epsilon} {\sim} \mathcal{N}(\mathbf{0}, \mathbf{I})$ 
\EndFor
\EndIf
\EndFor
\end{algorithmic}
\end{algorithm}

For all disentanglement figures in the main paper, we use the following parameters:
\begin{itemize}
    \item Appearance-only SRN chairs / PhotoShape Chairs: $\NumFreeCoordSteps=15$, $\NumRepaintStepsFeats=50$, $\NumRepaintsFeats=10$; Mean Runtime for a single generation (batch size 1) on a RTX 4090 GPU: 11.61 sec
    \item Appearance-only SRN cars: $\NumFreeCoordSteps=15$, $\NumRepaintStepsFeats=80$, $\NumRepaintsFeats=40$; Mean Runtime for a single generation (batch size 1) on a RTX 4090 GPU: 34.75 sec
    \item Shape-only SRN chairs / PhotoShape Chairs: $\NumFreeCoordSteps=50$, $\NumRepaintStepsFeats=100$, $\NumRepaintsFeats=2$; Mean Runtime for a single generation (batch size 1) on a RTX 4090 GPU: 9.15 sec
    \item Shape-only SRN cars: $\NumFreeCoordSteps=50$, $\NumRepaintStepsFeats=0$, $\NumRepaintsFeats=0$; Mean Runtime for a single generation (batch size 1) on a RTX 4090 GPU: 8.85 sec
\end{itemize}

\newpage

\section{Analysis}

As described in the main paper, we conduct ablations studies regarding the effects of different initialization strategies, feature dimensionality and regularization methods in the category-level Point-NeRF autodecoder and diffusion model. These parameters affect both, the quality of the reconstructions and the structure of the neural point cloud feature space. Both properties in turn affect the diffusion model that is trained on the resulting neural point clouds. As the reconstructed objects serve as training data for the diffusion model, the reconstruction quality likely is the upper bound of the generation quality. On the other hand, the structure of the feature space affects how well the data distribution can be learned by the diffusion model. 

\subsection{Setup} 
We conduct the analyses with similar settings as described in the main paper. The main difference is that, for computational reasons, we conduct the analyses with a smaller transformer denoiser network with 40M parameters. Thus, the numbers of this configuration might differ from the best configuration in the main paper. At the end of the analysis, we compare this 40M parameter model to the 300M parameter model from the main paper for the best initialization strategy, dimensionality, and regularization parameters. We conduct the analyses on ShapeNet Cars and Chairs. In case the results on Cars are very clear, we omit the corresponding experiments on Chairs.

\subsection{Neural point cloud initialization}
\label{sec:NeuralPointcloudInitializationAppendix}
\begin{figure}[h!]
        \centering
        \subfloat[]{\includegraphics[width=0.4\linewidth]{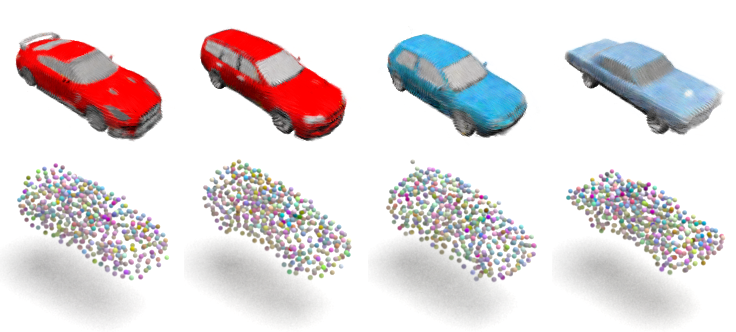}\label{subfig:InitializationAnalysesReconstructed-a} }\unskip\hspace*{0.01\linewidth}\vrule\hspace*{0.01\linewidth}
        \subfloat[]{
        \includegraphics[width=0.4\linewidth]{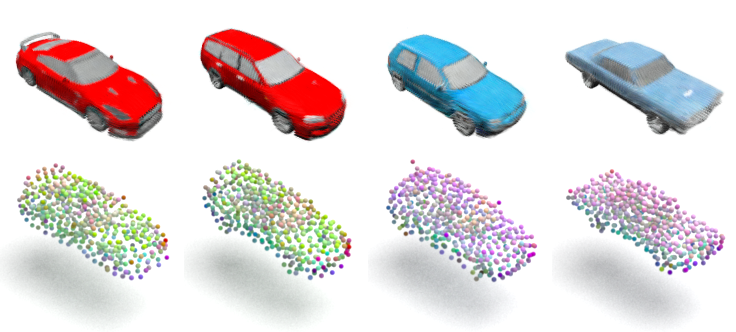}\label{subfig:InitializationAnalysesReconstructed-b} }
        \caption{The first row shows the Point-NeRF autodecoder renderings of four \emph{reconstructed} training objects. The second row shows a visualization of the features from the point clouds by taking the first three PCA components as RGB colors.  \protect\subref{subfig:InitializationAnalysesReconstructed-a} \textbf{Random initialization}: The features learned starting from a random initialization are distributed randomly across an object and differ across objects with the same appearance.  \protect\subref{subfig:InitializationAnalysesReconstructed-b} \textbf{Zero initialization}: Features learned starting with a zero initialization are coherent within an object and across objects with the same appearance.}
        \label{fig:InitializationAnalysesReconstructed}
\end{figure}

Regarding the initialization of the neural point cloud features, we analyze initialization with features sampled from a Gaussian distribution against a zero initialization. Interestingly, we find that these different initializations strongly affect the feature space of the trained models. To illustrate this, we visualize neural point features of reconstructed objects in~\cref{fig:InitializationAnalysesReconstructed}.

\begin{figure}[h!]
        \centering
        \subfloat[]{\includegraphics[width=0.4\linewidth]{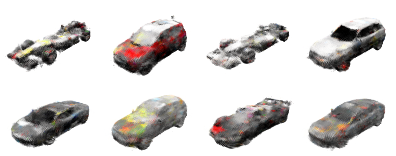}\label{subfig:InitializationAnalysesGenerated-a} }\unskip \vrule
        \subfloat[]{
        \includegraphics[width=0.4\linewidth]{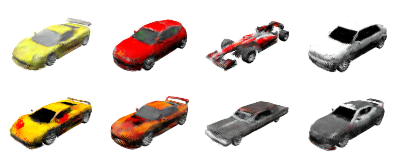}\label{subfig:InitializationAnalysesGenerated-b} }
        \caption{Generated samples from diffusion models trained on neural point clouds from category-level Point-NeRF autodecoders that were optimized with different initialization strategies: \protect\subref{subfig:InitializationAnalysesGenerated-a} \textbf{Random initialization} leads to many artifacts in the appearance of generated samples.  \protect\subref{subfig:InitializationAnalysesGenerated-b} \textbf{Zero initialization} leads to more diversity and fewer artifacts.}
        \label{fig:InitializationAnalysesGenerated}
\end{figure}

This effect is measured quantitatively via the cosine similarities in \cref{tab:Cosine} of the main paper. As shown in \cref{fig:InitializationAnalysesGenerated} and \cref{tab:Analyses}a, the more coherent features from the zero initialization, are vital to enable the successful training of a diffusion model.

\begin{table*}[t!]
\small
\centering
\setlength{\tabcolsep}{0.5mm}
\begin{tabular}{l
c
c
c
c|
>{\columncolor{bgcolor}}c >{\columncolor{bgcolor}}c >{\columncolor{bgcolor}}c c c
>{\columncolor{bgcolor}}c >{\columncolor{bgcolor}}c >{\columncolor{bgcolor}}c c c
}

\toprule
    \textbf{Setting}
    & \textbf{Dim.}
    & \textbf{Init.}
    & \textbf{Reg.}
    & $\lambda$
    & \multicolumn{5}{c}{\textbf{ShapeNet SRN Cars}}
    & \multicolumn{5}{c}{\textbf{ShapeNet SRN Chairs}}
    \\

    & 
    & 
    & 
    & 
    & PSNR$\uparrow$ & FID{\scriptsize rec}$\downarrow$ & KID{\scriptsize rec}$\downarrow$ & FID$\downarrow$ & KID$\downarrow$
    & PSNR$\uparrow$ & FID{\scriptsize rec}$\downarrow$ & KID{\scriptsize rec}$\downarrow$ & FID$\downarrow$ & KID$\downarrow$
    \\
    
\midrule

\textbf{a) Initialization}
& 
& 
& 
& 
& 
& 
& 
& 
& 
& 
& 
& 
& 
& 
\\

Random initialization
& 32
& Rand.
& \mn
& -
& 29.24 
& 37.24 
& 25.37 
& 125.51 
& 97.82 
& - 
& - 
& - 
& - 
& - 
\\

Zero initialization
& 32
& Zero
& \mn
& -
& 31.32 
& 18.96 
& 11.17 
& 53.55 
& 35.37 
& 34.91 
& 10.37 
& 4.85 
& 39.19 
& 23.64 
\\

\midrule

\textbf{b) Dimensionality}
& 
& 
& 
& 
& 
& 
& 
& 
& 
& 
& 
& 
& 
& 
\\

16D features
& 16
& Zero
& \mn
& -
& 30.60 
& 22.56 
& 13.86 
& 56.02 
& 39.30 
& - 
& - 
& - 
& - 
& - 
\\

32D features
& 32
& Zero
& \mn
& -
& 31.32 
& 18.96 
& 11.17 
& 53.55 
& 35.37 
& 34.91 
& 10.37 
& 4.85 
& 39.19 
& 23.64 
\\

128D features
& 128
& Zero
& \mn
& -
& 32.65 
& 19.33 
& 11.60 
& 73.93 
& 52.09 
& - 
& - 
& - 
& - 
& - 
\\

\midrule

\textbf{c) Regularization}
& 
& 
& 
& 
& 
& 
& 
& 
& 
& 
& 
& 
& 
& 
\\
    
No regularization
& 32
& Zero
& \mn
& -
& 31.32 
& 18.96 
& 11.17 
& 53.55 
& 35.37 
& 34.91 
& 10.37 
& 4.85 
& 39.19 
& 23.64 
\\

TV regularization
& 32
& Zero
& TV
& 3.5e-6
& 29.72 
& 22.42 
& 13.71 
& 45.90 
& 28.70 
& 32.38 
& 14.10 
& 6.70 
& 32.87 
& 17.49 
\\

KL regularization
& 32
& Zero
& KL
& 1e-6
& 30.02 
& 24.93 
& 15.60 
& 55.01 
& 35.86 
& 34.20 
& 8.37 
& 3.17 
& 18.13 
& 8.17 
\\

TV+KL regularization
& 32
& Zero
& TV,KL
& 3e-7,1e-7
& 29.70 
& 26.12 
& 16.44 
& 43.92 
& 26.53 
& 33.62 
& 8.58 
& 3.34 
& 17.17 
& 7.44 
\\

\midrule

\textbf{d) Model size}
& 
& 
& 
& 
& 
& 
& 
& 
& 
& 
& 
& 
& 
& 
\\

40M parameters
& 32
& Zero
& TV,KL
& 3e-7,1e-7
& 29.70 
& 26.12 
& 16.44 
& 43.92 
& 26.53 
& 33.62 
& 8.58 
& 3.34 
& 17.17 
& 7.44 
\\

300M parameters
& 32
& Zero
& TV,KL
& 3e-7,1e-7
& 29.70 
& 26.12 
& 16.44 
& 28.38
& 17.62
& 33.62 
& 8.58 
& 3.34 
& 9.87
& 3.62
\\
	
\bottomrule
\end{tabular}
\caption{\textbf{Analysis of the Point-NeRF autodecoder reconstructions and the diffusion model generations regarding}: \textbf{a)} feature initialization, \textbf{b)} feature dimensionality, \textbf{c)} feature regularization ($\lambda$ is the weight of the regularization loss) and \textbf{d)} model size. The PSNR, FID{\scriptsize rec} and KID{\scriptsize rec} metrics in the \colorbox{bgcolor}{gray columns} measure the quality of the reconstructions in the Point-NeRF autodecoder optimization stage. The FID and KID metrics measure the quality of the diffusion model generations. The reported KID is multiplied with $10^3$. 
Zero initialization clearly outperforms random initialization. 32D features outperform 16D and 128D features. Combined TV+KL regularization outperforms having no regularization or TV or KL regularization alone.
\label{tab:Analyses}
}
\end{table*}

\newpage

\subsection{Neural point cloud feature dimension}
\label{sec:NeuralPointcloudDimensionAppendix}

\begin{figure}[h!]
        \centering
        \subfloat[]{\includegraphics[width=0.30\linewidth]{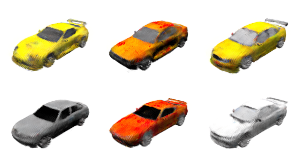}\label{subfig:DimensionAnalysisGenerated-a} }\unskip \vrule
        \subfloat[]{
        \includegraphics[width=0.30\linewidth]{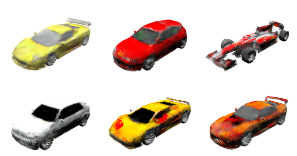}\label{subfig:DimensionAnalysisGenerated-b} }\unskip \vrule
        \subfloat[]{
        \includegraphics[width=0.30\linewidth]{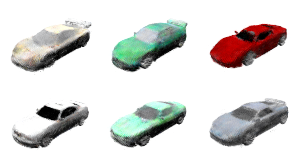}\label{subfig:DimensionAnalysisGenerated-c} }
        \caption{Generated samples from diffusion models trained on neural point clouds with different numbers of feature dimensions: \protect\subref{subfig:DimensionAnalysisGenerated-a} 16D features, \protect\subref{subfig:DimensionAnalysisGenerated-b} 32D features, \protect\subref{subfig:DimensionAnalysisGenerated-c} 128D features. We observe a clear difference between 16 and 32D  features, while the difference to 128D is small. 
        }
        \label{fig:DimensionAnalysisGenerated}
\end{figure}

We analyze 16D, 32D and 128D features for the neural point clouds in~\cref{tab:Analyses}b and~\cref{fig:DimensionAnalysisGenerated}. 

As indicated by the PSNR metrics, higher feature dimensions allow better training data reconstructions in the category-level Point-NeRF autodecoder training. However, the FID and KID metrics for the generation performance of the diffusion model decrease for 128D. 
As the visual quality of 32D and 128D in~\cref{fig:DimensionAnalysisGenerated} is very similar and the FID and KID metrics for 32D are better, we choose to  continue with the 32D features.

\subsection{Neural point cloud regularization}
\label{sec:NeuralPointcloudRegularizationAppendix}

\begin{figure}[h!]
    \def\tmpwidth{0.4\linewidth}
    \centering
    \subfloat[]{\includegraphics[width=\tmpwidth]{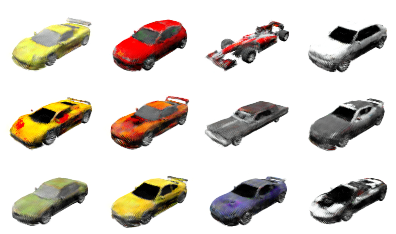}\label{subfig:RegularizationAnalysisGenerated-a} } \hspace*{0.07\linewidth}
    \subfloat[]{
    \includegraphics[width=\tmpwidth]{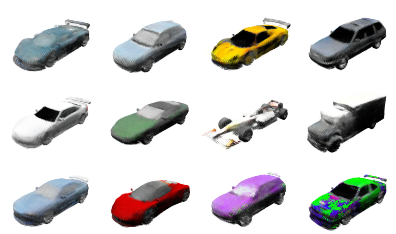}\label{subfig:RegularizationAnalysisGenerated-b} }\\\vspace*{0.03\linewidth}
    \subfloat[]{
    \includegraphics[width=\tmpwidth]{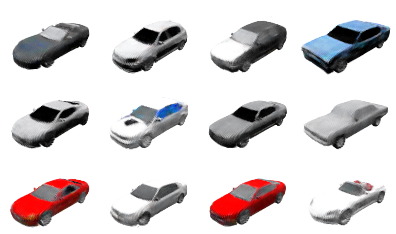}\label{subfig:RegularizationAnalysisGenerated-c} }\hspace*{0.07\linewidth}
    \subfloat[]{
    \includegraphics[width=\tmpwidth]{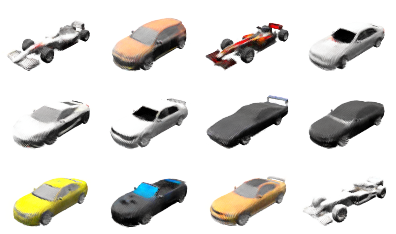}\label{subfig:RegularizationAnalysisGenerated-d} }
    \caption{Generated samples from diffusion models trained on neural point clouds from Point-NeRFs that were optimized with different regularization strategies: \protect\subref{subfig:RegularizationAnalysisGenerated-a} No regularization (FID 53.55), \protect\subref{subfig:RegularizationAnalysisGenerated-b} TV regularization (FID 45.90), \protect\subref{subfig:RegularizationAnalysisGenerated-c} KL regularization (FID 55.01), \protect\subref{subfig:RegularizationAnalysisGenerated-d} TV+KL regularization (FID 43.92). TV regularization increases performance regarding the FID and KID metrics. KL regularization leads to cleaner qualitative results. The proposed method \methodname{} uses a combination of both regularizations, which improves quantitative and qualitative results. 
    }
    \label{fig:RegularizationAnalysisGenerated}
\end{figure}

As stated in the main paper, we analyze the effects of applying KL and TV regularizations to the neural point features during the category-level Point-NeRF autodecoder training. 
The cosine similarities in \cref{tab:Cosine} of the main paper show that both regularizations further decrease the ambiguity of the latent space over the zero initialization. However, we observe different behaviors w.r.t. quantitative and qualitative results. On the one hand, among TV and KL regularization, TV leads to the better FID and KID scores in \cref{tab:Analyses}c. The qualitative comparison in \cref{fig:RegularizationAnalysisGenerated} on the other hand shows that KL regularization results in cleaner samples with less artifacts compared to TV regularization.
As a consequence, we try a combination of TV and KL regularization, which leads to an equally high visual quality and the best FID and KID scores. 

\subsection{Model size}

Lastly, we compare the performance of the transformer model with 40M parameters that was used in the analyses, with the performance of the transformer model with 300M parameters, that was used in the main paper. These models differ as follows:
\begin{itemize}
    \item 40M parameter model: 12 layers, hidden dimension 512, 8 heads; trained with batch size 64, learning rate $1e\!\!-\!\!4$, not using EMA on the model weights.
    \item 300M parameter model: 24 layers, hidden dimension 1024, 16 heads; trained with batch size 32, a learning rate $7e\!\!-\!\!5$ on Cars and $4e\!\!-\!\!5$ on Chairs, using EMA on the model weights. 
\end{itemize}

\paragraph{Quantitative Results.}
The quantitative comparison in \cref{tab:Analyses}d shows that the quality of the generated samples scales with the model size.

\begin{figure}[!h]
    \centering
      \def\tmpwidth{0.3\linewidth}
    \begin{tabular}{@{}c|c@{}|c@{}}
      \raisebox{0\height}{\rot{\small Gen}}\hspace{2mm}
      \includegraphics[width=\tmpwidth]{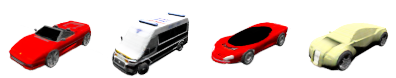} &
      \includegraphics[width=\tmpwidth]{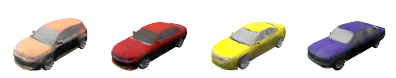} &
      \includegraphics[width=\tmpwidth]{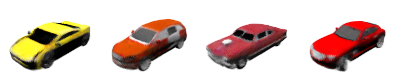} 
      \\
      \raisebox{-.1\height}{\rot{\small Train}}\hspace{2mm}
      \includegraphics[width=\tmpwidth]{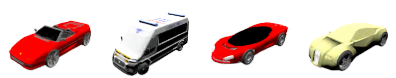} &
      \includegraphics[width=\tmpwidth]{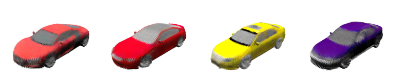} &
      \includegraphics[width=\tmpwidth]{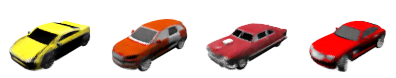} 
      \\
      SSDNeRF (FID: 11.08) & Ours, 40M model (FID: 43.92) & Ours, 300M model (FID: 28.38)
  \end{tabular}
  \caption{Analysis regarding similarity between generated objects and objects in the training set for SSDNeRF, our 40M parameter model, and our 300M parameter model. The top row shows generated samples and the bottom row shows the nearest neighbour sample in the training dataset. We retrieve the nearest neighbor training object based on the L2 distance between the Inception features of the generated and training images rendered from the same pose. One can observe that SSDNeRF and our 300M parameter model, which achieve good FID scores, generate samples that are very similar to specific samples in the training dataset.}
  \label{fig:nn_comparison}
\end{figure}

\paragraph{Overfitting.} We observed that 3D diffusion models on ShapeNet that achieve good FID scores, often produce samples that are very similar to specific samples in the training dataset. This is illustrated in \cref{fig:nn_comparison}, where we show generated samples and their nearest neighbour training sample for SSDNeRF~\cite{Chen2023}, our 40M parameter model, and our 300M parameter model.

Based on results from diffusion models for other representations (\eg Point-E for RGB point clouds, StableDiffusion for images), we expect that given larger datasets and appropriate conditioning mechanisms, the models scale well and are capable to generate diverse samples that are different from the training data. However, on small scale datasets like ShapeNet, our conclusion is that FID alone should be taken with a grain of salt.

\newpage
\paragraph{Overfitting and disentangled generation.}
Likewise, in disentangled generation, we observe that models that achieve good FID scores tend to be more ``locked-in''. For example in our appearance-only generation, the denoiser outputs are ignored for the point positions and instead their trajectory is forced via the forward diffusion process. We observe that this leads to stronger artifacts for models that achieve better FID scores. To resolve this, for these models, we use a higher number of resampling steps, which reduces the artifacts, but comes at the cost of longer runtimes. Further, we use a higher $\NumFreeCoordSteps$, \ie use more steps where the denoiser outputs are used to update the point positions via the reverse process. This reduces the artifacts, but comes at the cost of stronger deviations from the given input shape. 

\begin{figure}[h!]
        \centering
        \subfloat[]{
            \begin{tabular}{@{}cc@{}}
                \raisebox{.5\height}{\includegraphics[width=0.06\linewidth]{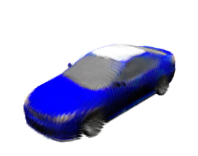}} &
              \includegraphics[width=0.4\linewidth]{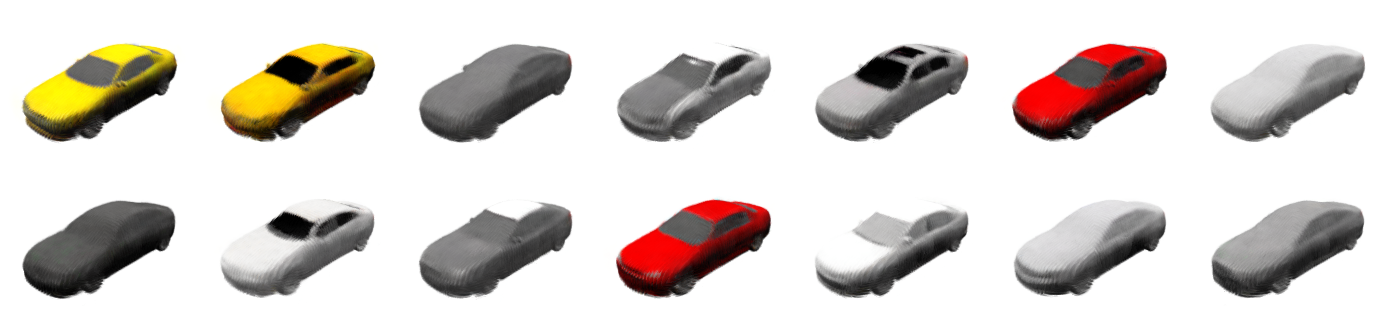} \\
            \raisebox{.5\height}{\includegraphics[width=0.06\linewidth]{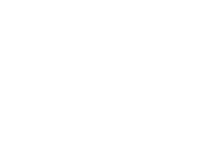}} &
              \includegraphics[width=0.4\linewidth]{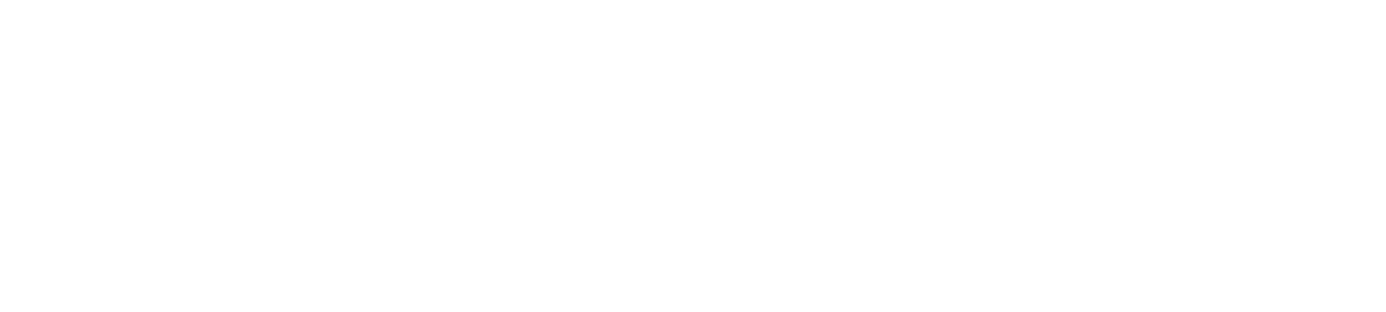} \\ 
                \raisebox{.5\height}{\includegraphics[width=0.06\linewidth]{figures/overfitting_disentanglement_analysis/white_base.png}} &
              \includegraphics[width=0.4\linewidth]{figures/overfitting_disentanglement_analysis/white_variations.png}
          \end{tabular}
          \label{subfig:Disentanglement40M}
        }\unskip \vrule
        \subfloat[]{
            \begin{tabular}{@{}cc@{}}
              \raisebox{.5\height}{\includegraphics[width=0.06\linewidth]{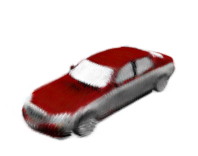}} &
              \includegraphics[width=0.4\linewidth]{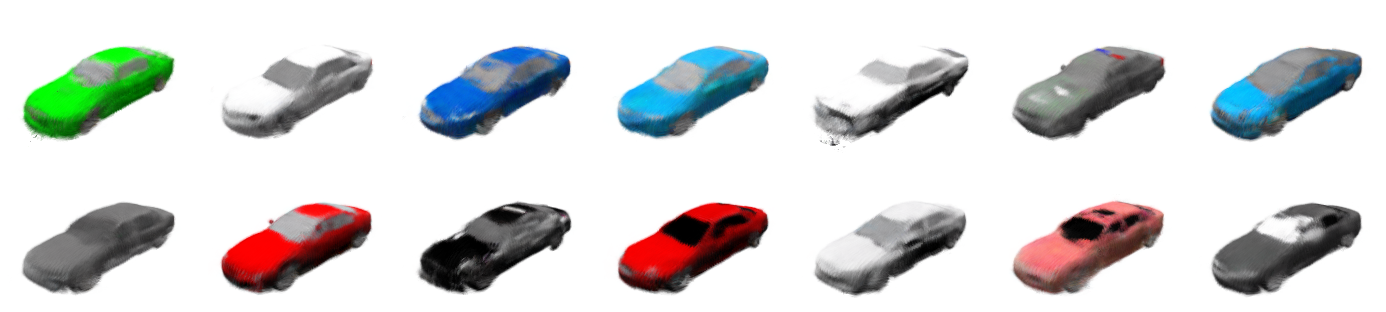} \\ \midrule
              \raisebox{.5\height}{\includegraphics[width=0.06\linewidth]{figures/overfitting_disentanglement_analysis/seed10_finalmodel_base.png}} &
              \includegraphics[width=0.4\linewidth]{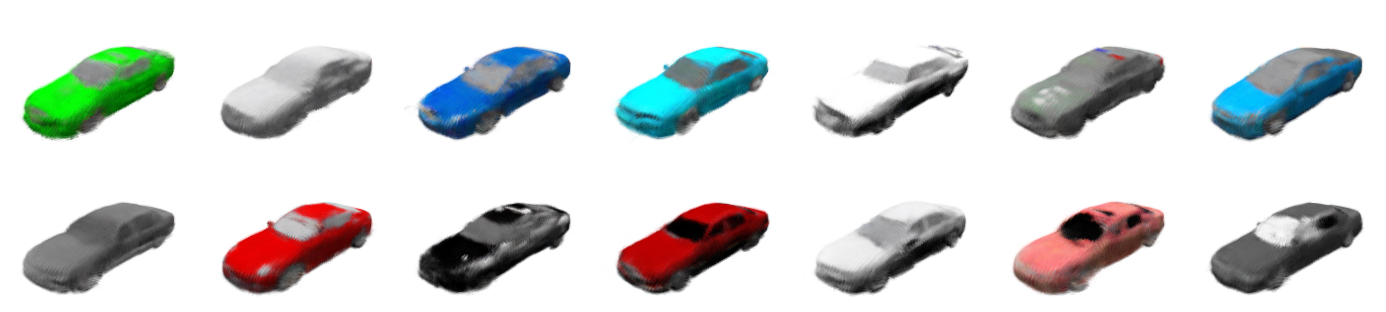} \\ \midrule
              \raisebox{.5\height}{\includegraphics[width=0.06\linewidth]{figures/overfitting_disentanglement_analysis/seed10_finalmodel_base.png}} &
              \includegraphics[width=0.4\linewidth]{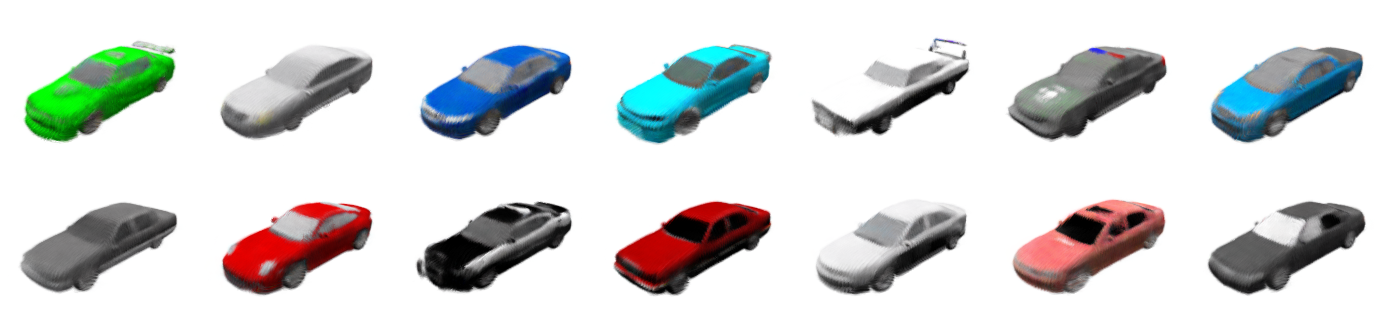} 
        \end{tabular}
        \label{subfig:Disentanglement300M}
        }
        \caption{\protect\subref{subfig:Disentanglement40M} Appearance-only generations on SRN cars for our 40M parameter model with parameters $\NumFreeCoordSteps=2$, $\NumRepaintStepsFeats=50$, $\NumRepaintsFeats=10$.  \protect\subref{subfig:Disentanglement300M} Appearance-only generations for our 300M parameter model. The top block uses the same parameters. One can observe that for this ``more locked-in model'', this results in more artifacts. 
        The middle block increases the number of resampling steps: $\NumFreeCoordSteps=2$, $\NumRepaintStepsFeats=80$, $\NumRepaintsFeats=40$. This reduces artifacts, but comes with longer runtimes. The bottom block increases the number of final steps $\NumFreeCoordSteps$ where the point position trajectory is not enforced via the forward process: $\NumFreeCoordSteps=15$, $\NumRepaintStepsFeats=80$, $\NumRepaintsFeats=40$. This basically resolves the artifacts, but the shape of the generated samples deviates more from the input shape.
        } 
        \label{fig:overfitting_disentanglement}
\end{figure}

We demonstrate this in \cref{fig:overfitting_disentanglement}, where we show non-cherrypicked appearance-only generations on SRN cars for our 40M parameter model and for our 300M parameter model with different disentangled generation parameters. We also observed that for SRN chairs and PhotoShape chairs, the effect is less pronounced.

\newpage

\section{Qualitative results for unconditional generation on ShapeNet Cars}
\begin{figure*}[h!]
        \centering
        \includegraphics[width=0.88\linewidth]{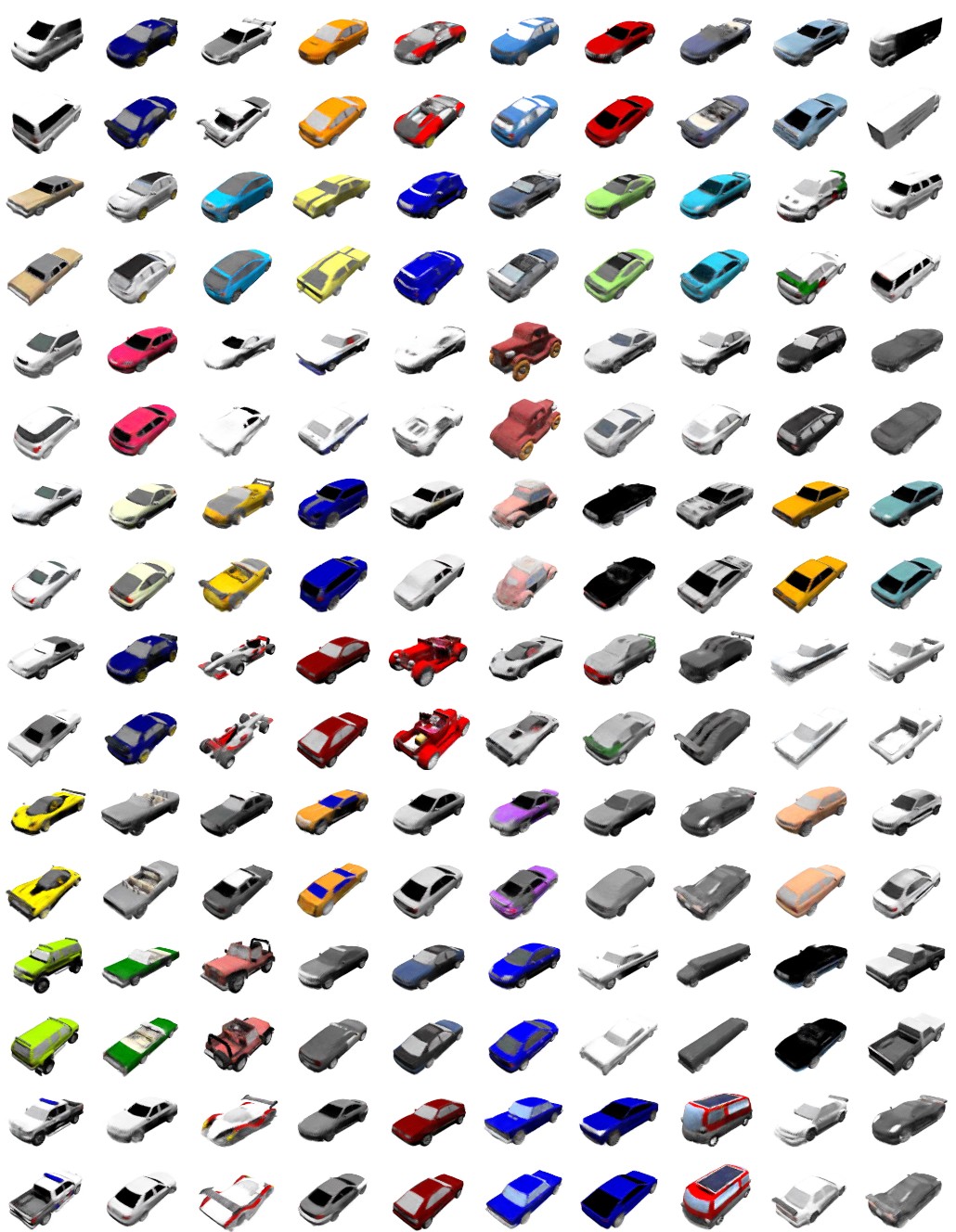}
        \caption{Unconditional generations from the proposed \methodname{} model trained on ShapeNet Cars. Each generated object is visualized from two different viewpoints. Note that the shown objects are not cherry-picked.}
        \label{fig:UnconditionalGenerationsCars}  
\end{figure*}

Figure~\ref{fig:UnconditionalGenerationsCars} shows unconditional generations of the proposed \methodname{} model trained on ShapeNet Cars.

\section{Qualitative results for unconditional generation on ShapeNet Chairs}
\begin{figure*}[h!]
        \centering
        \includegraphics[width=0.76\linewidth]{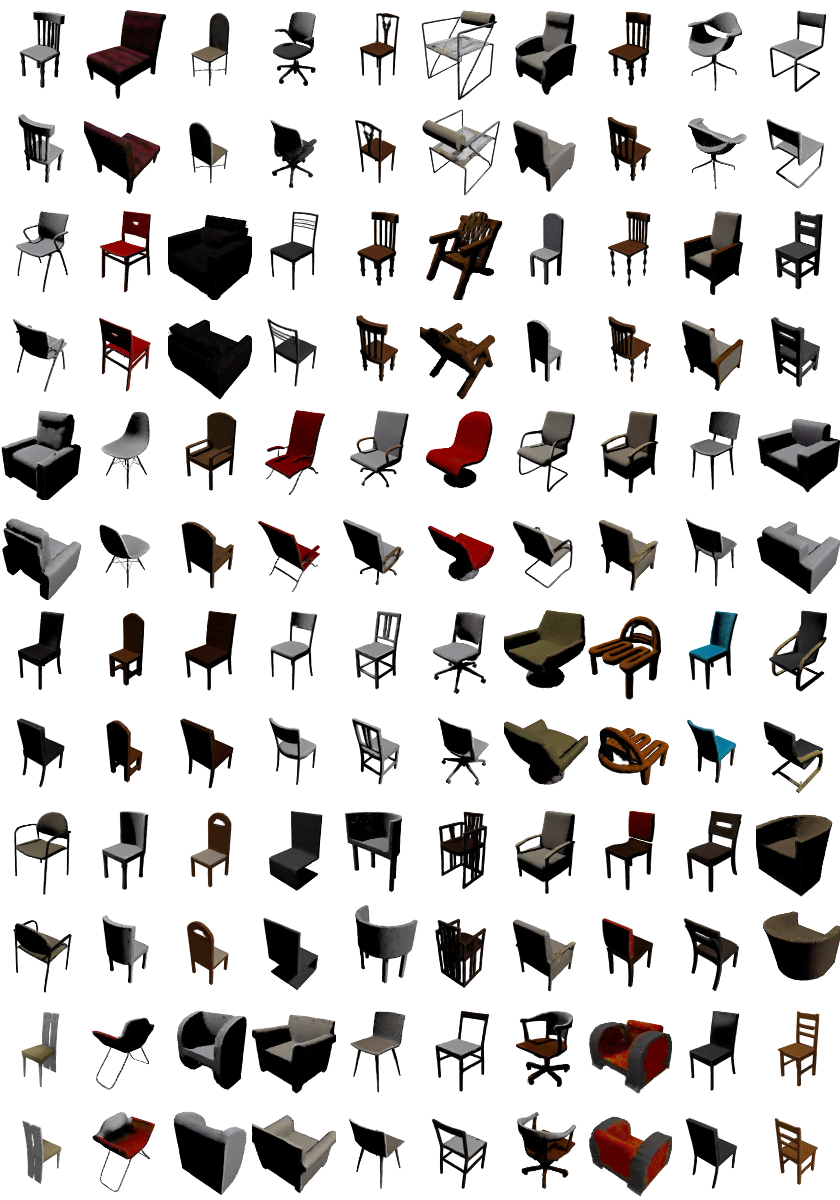}
        \caption{Unconditional generations of the proposed \methodname{} model trained on ShapeNet Chairs. Each generated object is visualized from two different viewpoints. Note that the shown objects are not cherry-picked.}
        \label{fig:UnconditionalGenerationsChairs}  
\end{figure*}

Figure~\ref{fig:UnconditionalGenerationsChairs} shows unconditional generations from the proposed \methodname{} model trained on ShapeNet Chairs.

\section{Baseline details}

For our comparisons on ShapeNet Cars, ShapeNet Chairs, and PhotoShape Chairs, we retrain GRAF and Disentangled3D on these datasets.
For training, both approaches require a dataset of images, the distribution of camera poses $p_\xi$ that correspond to the images, and a known camera matrix $\mathbf{K}$. Further, the volumetric rendering in both approaches requires given near and far clipping planes. Training is conducted in a GAN framework. The generator renders images for randomly sampled shape and appearance latent codes and randomly sampled camera poses. The discriminator compares generated and real images. 

\paragraph{ShapeNet Cars.}
For training on ShapeNet Cars, according to the the SRN rendering parameters, we set the radius to $1.3$, the field of view to $52$ (chosen such that it results in the correct focal length), and sample camera poses from the full hemisphere. To define the near and far planes, we compute the cube that bounds all pointclouds. For the Cars training split, this gives:
\begin{itemize}
    \item x-coordinate range: $-0.30903995$ to $0.30898672$
    \item y-coordinate range: $-0.48859358$ to $0.49043328$
    \item z-coordinate range: $-0.27073205$ to $0.2709335$
\end{itemize}

The nearest possible point hence has the following distance to the camera: \newline$1.3 - \sqrt{0.30903995^2 + 0.49043328^2 + 0.2709335^2} = 1.3 - 0.6398714 = 0.6601285815238953$.

The furthest point hence has the following distance to the camera: $1.3 + 0.6398714 = 1.9398714184761048$.

Based on this, we set the near and far planes to $0.5$ and $2.1$.

\paragraph{ShapeNet Chairs.} For training on ShapeNet Cars, according to the the SRN rendering parameters, we set the radius to $2.0$, the field of view to $52$ (chosen such that it results in the correct focal length), and sample camera poses from the full hemisphere. To define the near and far planes, we compute the cube that bounds all pointclouds. For the Chairs training split, this gives:
\begin{itemize}
    \item x-coordinate range: $-0.5$ to $0.5$
    \item x-coordinate range: $-0.5$ to $0.5$
    \item x-coordinate range: $-0.5$ to $0.5$
\end{itemize}

The nearest possible point hence has the following distance to the camera: $2.0 - \sqrt{0.5^2 + 0.5^2 + 0.5^2} = 2.0 - 0.87 = 1.13$.

The furthest point hence has the following distance to the camera: $2. + 0.87 = 2.87$.

Based on this, we set the near and far planes to $1.0$ and $3.0$.

\paragraph{PhotoShape Chairs.} For training on PhotoShape Chairs, according to the PhotoShape Chairs rendering parameters, we set the radius to $2.5$ and the field of view to $39.6$ (chosen such that it results in the correct focal length). As PhotoShape Chairs views were always rendered from the same set of camera poses, we randomly sample poses from this discrete set during training. As the 3D object shapes are the same as on ShapeNet Chairs, we compute the near and far clipping planes comparably to ShapeNet Chairs. Given the different radius, this results in near and far planes of $1.25$ and $3.75$.

\newpage
\section{Comparison to a Combination of Shape and Texture Generation}
Recently, multiple works applied diffusion models for 3D shape generation~\cite{zheng2023lasdiffusion,zhang20233dshape2vecset,shim2023diffusion,li2023diffusionsdf,Shue_2023_CVPR} and for the generation of textures for a given 3D mesh~\cite{chen2023text2tex,youwang2023paintit,cao2023texfusion}. Combining the two approaches, allows generating a shape, followed by generation of different appearances. For completeness, in the following, we conduct a comparison to such a combination. However, please note that separate 3D shape and texture generation differs strongly in concept from our work and has different assumptions and results: \textbf{(1)} with separate generation it is only possible to re-generate the appearance for a given shape, but not vice-versa. \textbf{(2)} SOTA texture generation approaches use test-time optimization with score distillation from a 2D diffusion model, which results in generation times in the order of minutes (we measure 9 minutes for Text2Tex), where our method takes only a few seconds to generate a sample and is thus orders of magnitude faster. \textbf{(3)} Both approaches differ strongly in terms of training data: our model uses a rather small dataset of 3D objects (a few thousand), whereas Text2Tex builds on a diffusion model trained on large scale image data (on the order of billions). 

Despite these fundamental differences, we provide a comparison between our method and the mentioned ``Locally Attentional SDF Diffusion'' plus ``Text2Tex'' models (running their code as it is on Github)  below. The following shows four resulting appearance-only generations for text prompts ``\texttt{a <black/red/green/blue> chair}'' in comparison to a sample from our method with resampled appearance:

\begin{figure}[!h]
  \vspace{-0.1cm}
      \centering
      \def\tmpwidth{0.1\linewidth}
    \begin{tabular}{ccccc}
      \rotatebox{90}{\parbox{1.2cm}{\small LAS$\rightarrow$T2T\\\phantom{x}(9 min)}} &
      \includegraphics[width=\tmpwidth]{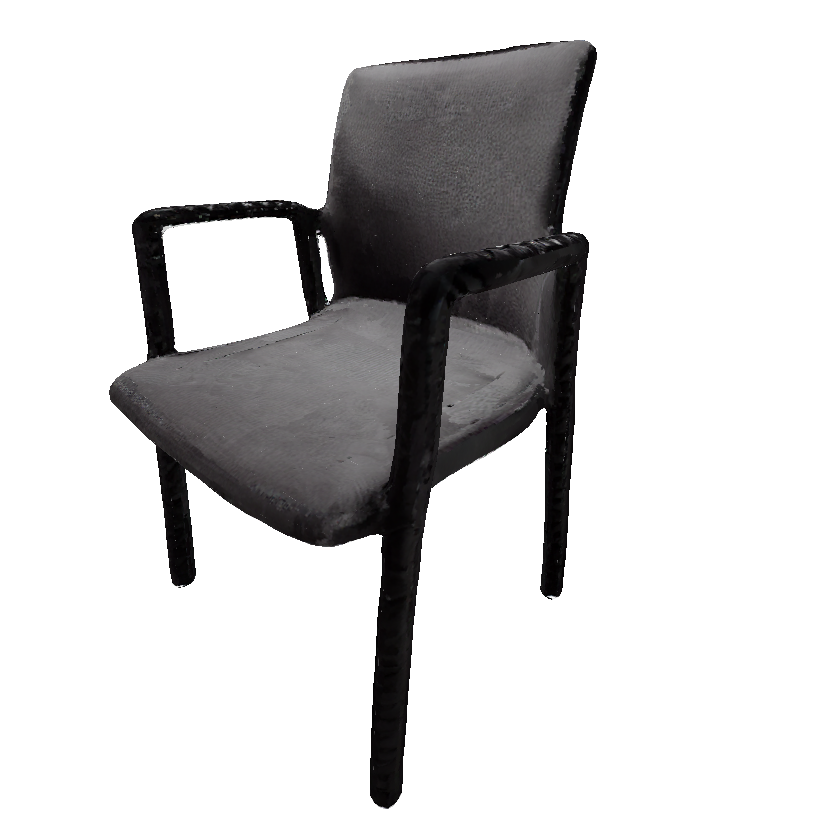} &
      \includegraphics[width=\tmpwidth]{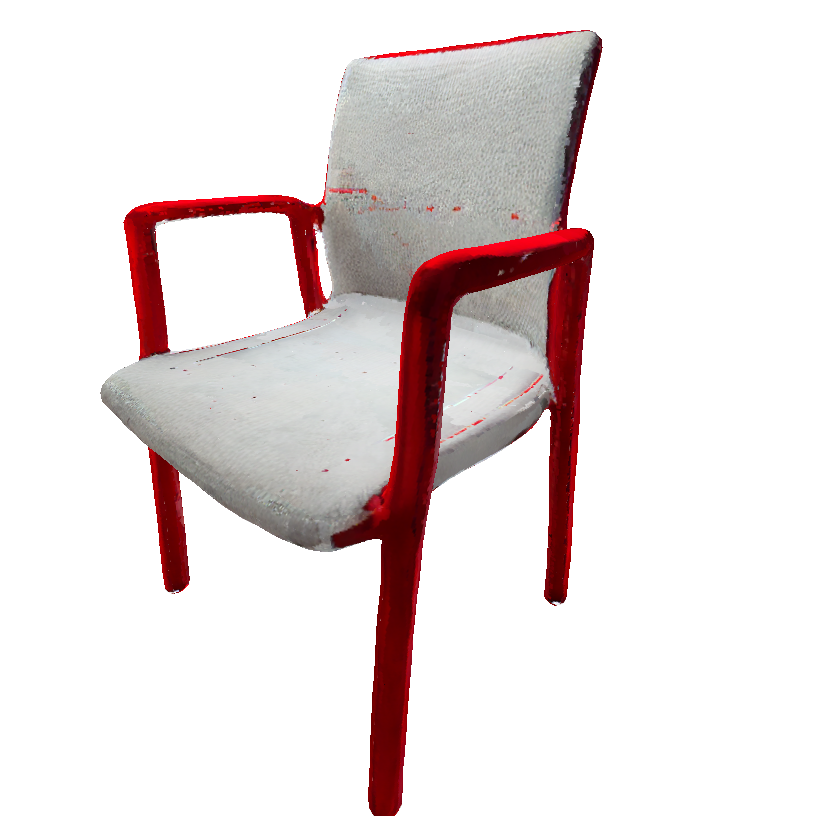} &
      \includegraphics[width=\tmpwidth]{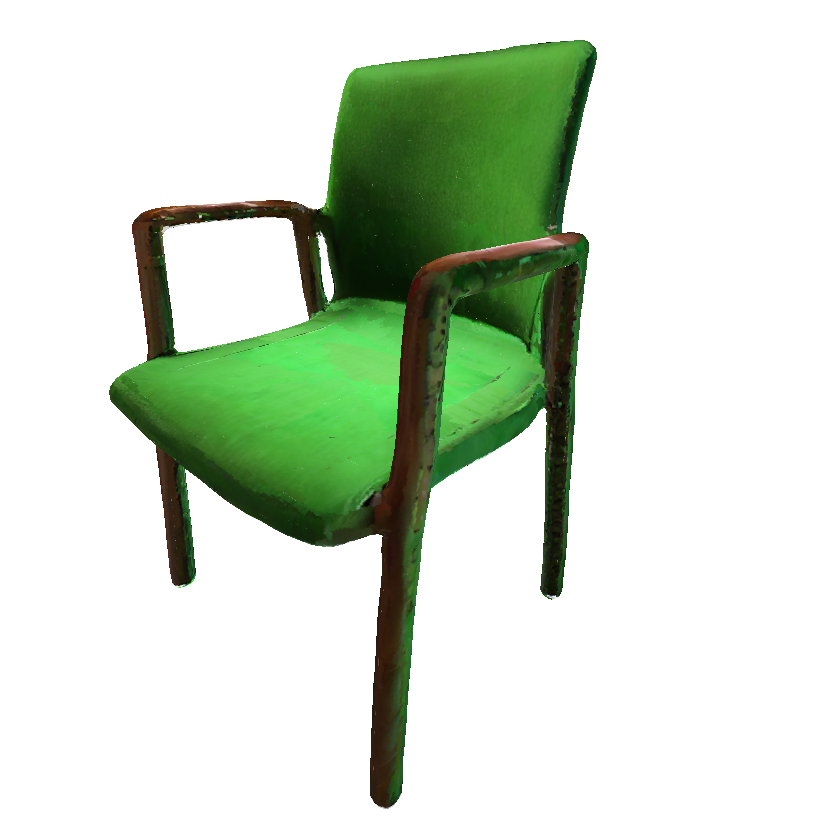} &
      \includegraphics[width=\tmpwidth]{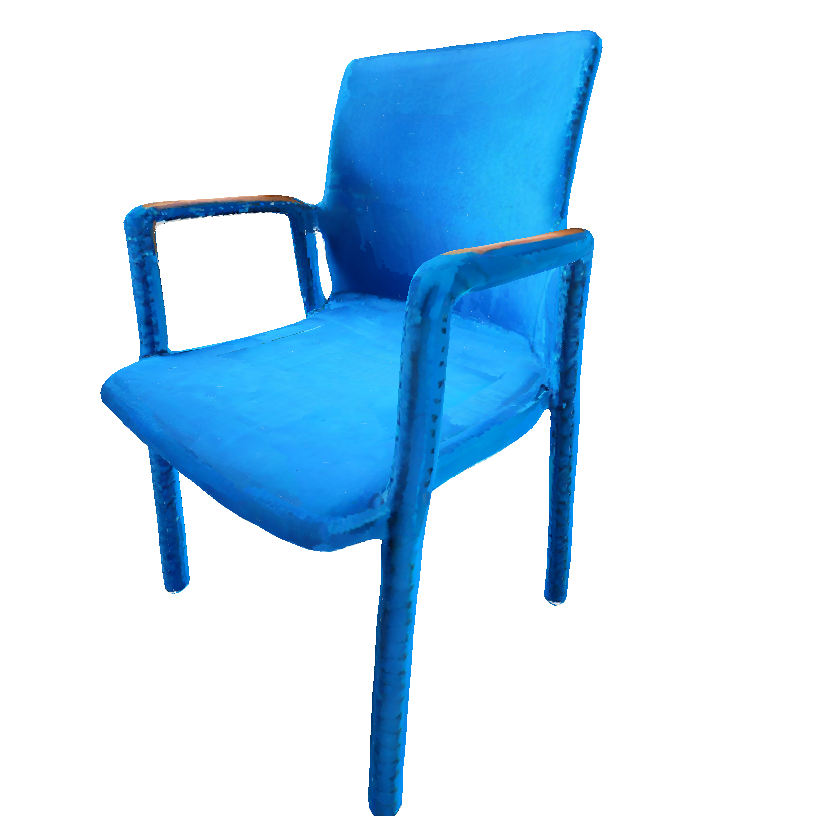} 
      \\
      \midrule
      \rotatebox{90}{\parbox{1.6cm}{\small $\quad$Ours\\(11.6 sec)}} &
      \includegraphics[width=\tmpwidth]{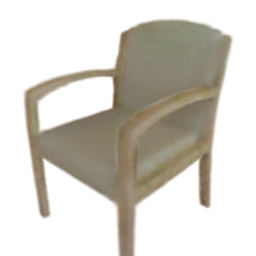} &
      \includegraphics[width=\tmpwidth]{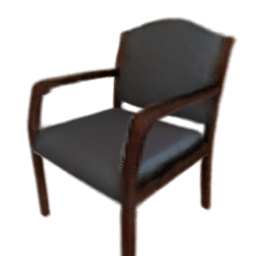} &
      \includegraphics[width=\tmpwidth]{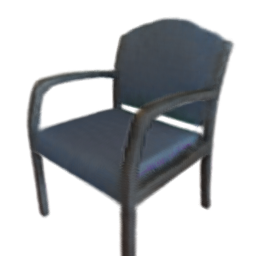} &
      \includegraphics[width=\tmpwidth]{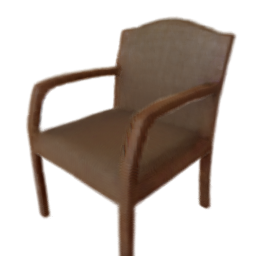} 
  \end{tabular}
  \caption{Comparison of our model and a combination of Locally Attentional SDF Diffusion~\cite{zheng2023lasdiffusion} plus Text2Tex~\cite{chen2023text2tex}.}
\end{figure}

It can be seen that our method finds a good trade-off between quality and generation time.
In general, we think that both, generative modeling on 3D representations and generative modeling on 2D images plus score distillation, are interesting research directions, but have different advantages and disadvantages. 

\end{document}